\documentclass{article}

\usepackage{arxiv}

\usepackage[utf8]{inputenc} 
\usepackage[T1]{fontenc}    
\usepackage{hyperref}       
\usepackage{url}            
\usepackage{booktabs}       
\usepackage{amsfonts}       
\usepackage{nicefrac}       
\usepackage{microtype}      
\usepackage{lipsum}
\usepackage{graphicx}
\graphicspath{ {./images/} }

\usepackage{microtype}
\usepackage{graphicx}
\usepackage{subfigure}
\usepackage{booktabs} 
\newcommand{\rom}[1]{\romannumeral #1\relax}

\usepackage{amsmath}
\usepackage{amssymb}
\usepackage{mathtools}
\usepackage{amsthm}
\usepackage{bm}
\usepackage{enumitem}
\usepackage{setspace}
\usepackage{color}
\usepackage{natbib}
\usepackage[capitalize,noabbrev]{cleveref}
\usepackage{xcolor}
\usepackage{wrapfig}

\definecolor{myblue}{HTML}{0036A2}

\title{StarMAP: Global Neighbor Embedding for \\ Faithful Data Visualization}

\author{
 Koshi Watanabe \\
  Hokkaido University\\
  \texttt{koshi@lmd.ist.hokudai.ac.jp} \\
   \And
 Keisuke Maeda \\
  Hokkaido University \\
  \texttt{maeda@lmd.ist.hokudai.ac.jp} \\
  \And
 Takahiro Ogawa \\
  Hokkaido University\\
  \texttt{ogawa@lmd.ist.hokudai.ac.jp} \\
  \And
 Miki Haseyama \\
  Hokkaido University\\
  \texttt{mhaseyama@lmd.ist.hokudai.ac.jp} \\
}

\begin{document}
\maketitle
\begin{abstract}
Neighbor embedding is widely employed to visualize high-dimensional data; however, it frequently overlooks the global structure, e.g., intercluster similarities, thereby impeding accurate visualization. To address this problem, this paper presents \underline{Star}-attracted \underline{M}anifold \underline{A}pproximation and \underline{P}rojection (StarMAP), which incorporates the advantage of principal component analysis (PCA) in neighbor embedding. Inspired by the property of PCA embedding, which can be viewed as the largest shadow of the data, StarMAP introduces the concept of \textit{star attraction} by leveraging the PCA embedding. This approach yields faithful global structure preservation while maintaining the interpretability and computational efficiency of neighbor embedding. StarMAP was compared with existing methods in the visualization tasks of toy datasets, single-cell RNA sequencing data, and deep representation. The experimental results show that StarMAP is simple but effective in realizing faithful visualizations.
\end{abstract}


\section{Introduction}
\label{sec:introduction}
Real-world data are represented as high-dimensional vectors, which makes the data difficult to interpret. In addition, high-dimensional space causes the \textit{curse of dimensionality}, which degrades the performance of many algorithms. Dimensionality reduction (DR)~\cite{maaten2009dimensionality} solves these problems by deriving low-dimensional embeddings from the data. In particular, DR into a two-dimensional space facilitates \textit{data visualization}~\cite{rudin2022interpretable}, which helps us to identify underlying patterns in the data. Data visualization has been widely studied to enable an intuitive analysis of complex data, e.g., deep representation or single-cell RNA sequencing data. 

\par
DR methods have advanced over the years. A fundamental approach involves spectral-based reduction through eigenvalue decomposition. Principal component analysis (PCA)~\cite{hotelling1933analysis} is a classical method that assumes that the dominant eigenvectors of the covariance matrix encapsulate most of the information, reasonably approximating the original data. 
However, PCA does not consider the local structure, frequently resulting in crowded embedding, which is unsuitable for visualization. To address this limitation, neighbor embedding assumes that local information is informative and seeks to minimize the loss of neighbor relations. t-stochastic neighbor embedding (t-SNE)~\cite{van2008visualizing, van2014accelerating} is considered a gold standard because its embedding captures the local structure effectively, particularly the cluster structure. Uniform manifold approximation and projection (UMAP)~\cite{mcinnes2018umap} has emerged as another standard that enhances the scalability of neighbor embedding. Now, t-SNE and UMAP serve as baselines in data visualization, with recent studies exploring their relation~\citep{damrich2022t, bohm2022attraction, jenssen2024map} or proposing improvement~\citep{wang2021understanding,bohm2023unsupervised, li2024fedne}.
\par
A significant issue with neighbor embedding is that it frequently overlooks the global structure of the data~\cite{wang2021understanding, kobak2021initialization}. For example, a cell population is the local structure, 
\begin{wrapfigure}{r}{0.6\textwidth}
    \centering
    {
        \includegraphics[width=95mm]{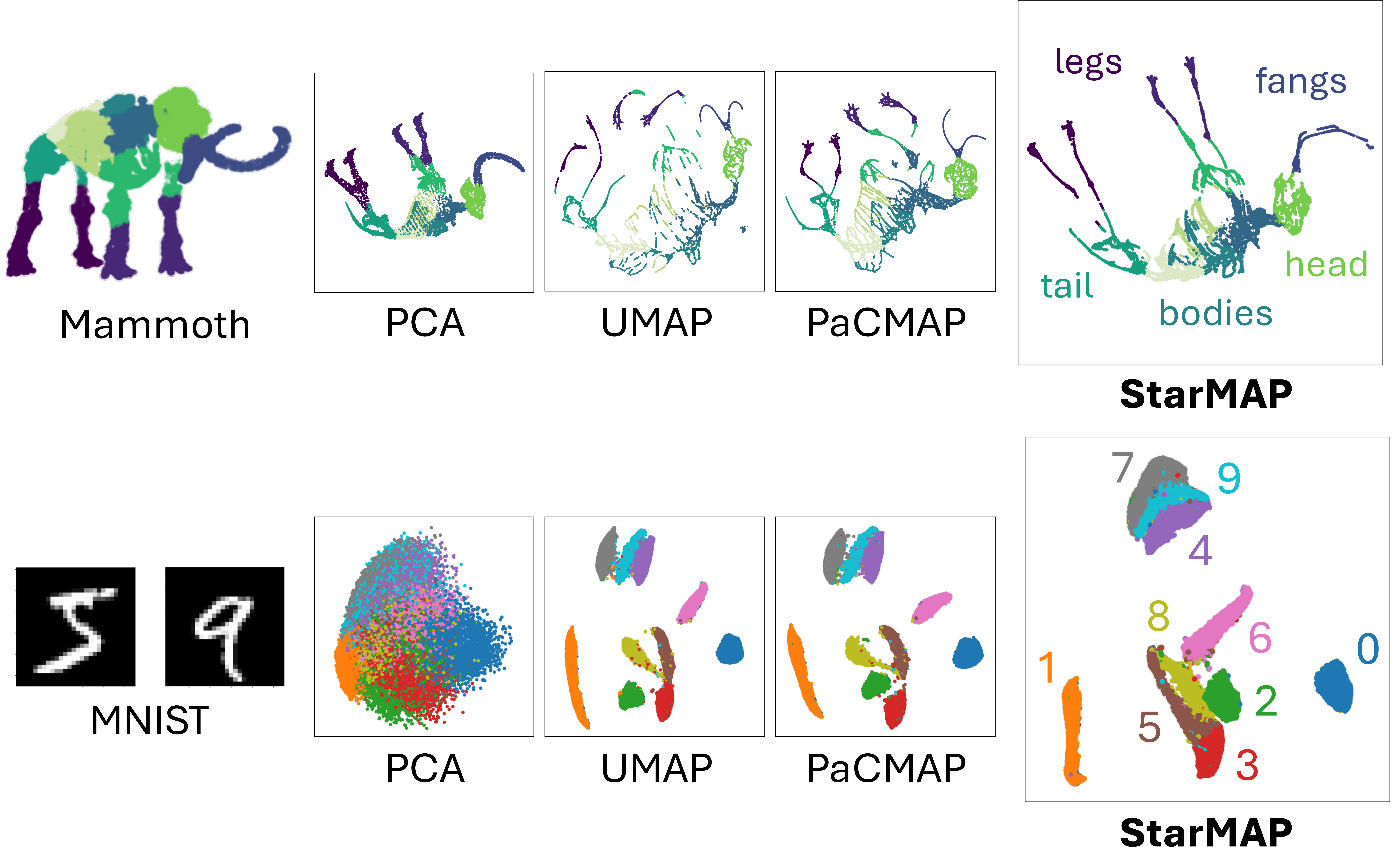}
    }
    \caption{Visualization results on the Mammoth and MNIST datasets with PCA, UMAP, PaCMAP, and StarMAP (\textbf{ours}).}
    \label{fig:abstract}
\end{wrapfigure}
and the intercell similarity is the global structure of the data. The neighbor embedding methods focus solely on neighbor relations; thus, the absence of the global structure is inevitable. 
However, \textit{faithful} data visualization, which preserves both the global and local structures, is crucial for uncovering meaningful patterns, e.g., a cellular lineage in developmental biology or cooccurrence relationships among identified clusters.
\par
Previous methods modify the neighbor estimation. For example, TriMAP~\cite{amid2019trimap} introduces the triplet relation, PHATE~\cite{moon2019visualizing} employs the diffusion operation to capture the multihop neighbor relation, and PaCMAP~\cite{wang2021understanding} focuses on the similarity between the intermediate neighbor points. However, experimental results indicate that while these improvements are promising, they remain limited, and global neighbor embedding is still being investigated.
\par
Another approach modifies the initialization. The neighbor embedding depends on gradient-based optimization; thus, an initial embedding is required. Kobak et al.~\cite{kobak2021initialization} highlighted that PCA initialization helps preserve the global structure in single-cell RNA sequencing data. One explanation for this phenomenon is that PCA maximizes the variance of the embedding within a subspace under a linearity constraint, thereby retaining as much global information as possible in the resulting embedding. The PCA embedding is \textit{the largest shadow} within the subspace. Consequently, one hypothesis naturally arises: the PCA embedding is the most accurate concerning the global structure. However, recent research~\cite{kobak2019art,kobak2021initialization} has adhered to the initialization-based approaches, which do not ensure the preservation of the initial PCA embedding after optimization.
\par
This paper presents a novel neighbor embedding method, \textit{Star-attracted Manifold Approximation and Projection (StarMAP)}, to achieve faithful data visualization that preserves both the global and local structures. StarMAP posits that the PCA embedding offers the most accurate visualization of the data concerning the global structure, and we incorporate this advantage into UMAP. StarMAP is closely related to the attraction-repulsion framework of neighbor embedding methods~\cite{bohm2022attraction, damrich2022t, jenssen2024map} which has been studied to uncover relationships among neighbor embedding algorithms. Based on the description of this framework, UMAP seeks to achieve an equilibrium between the attraction of neighboring points and the repulsion of randomly sampled points. StarMAP introduces a novel attraction force, i.e., \textit{star attraction}, to preserve the PCA embedding effectively. Initially, StarMAP computes anchor points for the high-dimensional data along with an assignment function, then simultaneously embeds both the data and the anchors into the low-dimensional space using PCA. We refer to the low-dimensional embedding of the anchor points as \textit{stars} and keep them fixed during optimization while applying their attraction on the assigned points. StarMAP explicitly preserves the global structure via the star attraction, a feature that was challenging for previous approaches to achieve. 
\par
Figure~\ref{fig:abstract} compares the embeddings of PCA, UMAP, PaCMAP, and StarMAP on the Mammoth and MNIST datasets, which are commonly used to validate data visualization methods. The key structure in the Mammoth dataset is the interbody part relationship, while in the MNIST dataset, it is the digit clusters. As mentioned previously, PCA effectively preserves the \textit{global} shape of the mammoth, and UMAP captures the \textit{local} digit clusters. However, both methods struggle to preserve the structure of both datasets. In contrast, StarMAP successfully preserves the shape of the mammoth while identifying the digit clusters in MNIST, thereby demonstrating its efficacy from a visual perspective.
\par
\textbf{The primary contribution of this paper} is the development of StarMAP to visualize the global and local structures of data, which is challenging for previous methods, including t-SNE and UMAP. The central innovation is \textit{star attraction}, which integrates the advantage of PCA into neighbor embedding. The experimental results demonstrate that StarMAP is a straightforward yet highly effective technique to faithfully visualize high-dimensional data, including single-cell RNA sequencing data and deep representation.

\section{Related Work}
\label{sec:related_works}
\subsection{Spectral-Based Methods}
\label{subsec:spectral_methods}
Spectral methods are grounded in the spectral theory and derive the low-dimensional embedding through eigenvalue decomposition. PCA~\cite{hotelling1933analysis} decomposes the covariance matrix, and its embedding maximizes the variance within the subspace, which is considered to be the largest shadow of the data. Multidimensional scaling (MDS)~\cite{torgerson1958theory} provides a general framework that utilizes a distance matrix to derive the low-dimensional embedding, and Isomap~\cite{balasubramanian2002isomap} is a variant of MDS that utilizes the geodesic distance. However, these methods do not consider the local structure, which results in crowded visualizations. Laplacian eigenmaps~\cite{belkin2003laplacian} serve as an intermediate method by decomposing the Laplacian of the neighborhood graph where the neighborhood graph is estimated based on the point-wise distance. The relationship between the Laplacian Eigenmaps and the neighbor embedding methods is discussed in the literature~\cite{bohm2022attraction}.

\subsection{Neighbor Embedding Methods}
\label{subsec:ne_methods}
Neighbor embedding methods attempt to minimize the loss of neighbor relations. t-SNE~\cite{van2008visualizing} is considered a gold standard that employs KL divergence as the loss function. The t-SNE embedding is much more discriminative than previous spectral-based methods; thus it is particularly effective for data visualization. Numerous variants of t-SNE have been developed, e.g., parametric~\citep{van2009learning}, triplet~\citep{van2012stochastic}, and computationally efficient extensions~\cite{van2014accelerating, fu2019atsne}. However, KL divergence requires normalization of the neighborhood graph, which complicates scalable optimization. UMAP~\cite{mcinnes2018umap} enhances the scalability of neighbor embedding algorithms by leveraging a fuzzy set cross-entropy loss that does not require normalization, similar to that of LargeVis~\cite{tang2016visualizing}, thereby reducing the computational complexity. Both t-SNE and UMAP have become standard baselines for DR-based data visualization. 
\par
Recent studies have investigated the relationship between neighbor embedding methods and their improvement. The attraction-repulsion framework~\cite{bohm2022attraction, jenssen2024map} and contrastive learning-based analysis~\cite{artemenkov2020ncvis, damrich2022t} are valuable for uncovering their relations. The major problem with neighbor embedding methods is their inability to preserve the global structure, which has been considered in efforts for improvement. For example, PaCMAP~\cite{wang2021understanding} introduces the middle neighbor similarity to preserve the global structure effectively. In addition, initialization-based improvements, particularly using PCA initialization, also play a crucial role~\cite{kobak2019art, kobak2021initialization}. However, while these improvements show promise, they remain limited. We incorporate the PCA embedding into the neighbor embedding to address this limitation.

\section{UMAP: An Overview}
\label{sec:umap}
The formal definition of DR involves estimating the low-dimensional embedding, denoted $\bm{\mathrm{Y}}=[\bm{\mathrm{y}}_{1},\,\bm{\mathrm{y}}_{2},\,\ldots,\bm{\mathrm{y}}_{N}]^{\top}\in\mathbb{R}^{N \times Q}$, from the high-dimensional observation $\bm{\mathrm{X}}=[\bm{\mathrm{x}}_{1},\,\bm{\mathrm{x}}_{2},\,\ldots,\bm{\mathrm{x}}_{N}]^{\top}\in\mathbb{R}^{N \times D}$, where $Q$ is the number of dimensions of the low-dimensional space, $D$ is that of the observed space ($D \gg Q$), and $N$ is the number of samples. 
UMAP constructs its embedding by estimating the $k$ nearest neighbors ($k$NNs), justified from the principles of algebraic topology theory~\cite{may1992simplicial}. The $(i,j)$-th entry, i.e., $w_{ij}$, of the weighted adjacency matrix $\bm{\mathrm{W}}$ is defined as follows:
\begin{align}
    w_{ij}&=w_{j|i} + w_{i|j}- w_{j|i} \cdot w_{i|j},\label{eq:wij}\\
    w_{j|i}&=
      \begin{cases}
        \exp{\left[-\frac{\mathrm{d}(\bm{\mathrm{x}}_{i},\bm{\mathrm{x}}_{j})-\rho_{i}}{\sigma_{i}}\right]} & (j\in\{i_{1},\ldots,i_{k}\}), \\
        0 & (\mathrm{otherwise}),
      \end{cases}\label{eq:wji}
\end{align} 
where $i_{l}\,(l=1,2,\ldots,k)$ is an index of the $l$-nearest neighbor of $\bm{\mathrm{x}}_{i}$ measured by a distance function $\mathrm{d}(\cdot,\cdot)$ (generally Euclidean distance), $\rho_{i}=\mathrm{d}(\bm{\mathrm{x}}_{i},\bm{\mathrm{x}}_{i_{1}})$ is the distance from the nearest neighbor, and $\sigma_{i}$ is a hyperparameter indicating the local connectivity around $\bm{\mathrm{x}}_{i}$. 
\par
UMAP optimizes the low-dimensional embedding $\bm{\mathrm{Y}}$ based on the weighted adjacency matrix $\bm{\mathrm{W}}$. Before optimization, UMAP initializes $\bm{\mathrm{Y}}$ using either Laplacian eigenmaps or PCA and subsequently normalizes its scale. The loss function is the fuzzy set cross-entropy as follows:
\begin{align}
    \mathcal{L}&=-\sum_{i,j} w_{ij}\log v_{ij} - \sum_{i,j} (1-w_{ij})\log (1-v_{ij}),\label{eq:obj}
\end{align}
where $ v_{ij}=\left(1+a||\bm{\mathrm{y}}_{i}-\bm{\mathrm{y}}_{j}||_{2}^{2b}\right)^{-1}$ represents the heavy-tailed similarity between $\bm{\mathrm{y}}_{i}$ and $\bm{\mathrm{y}}_{j}$ with hyperparameters $a$ and $b$. Typically, stochastic gradient descent is employed to minimize the loss function in Eq.~\eqref{eq:obj}. The gradient of this loss can be interpreted as the sum of the attraction and repulsion forces~\cite{mcinnes2018umap,bohm2022attraction}, with UMAP seeking the equilibrium between these forces. Formally, each point $\bm{\mathrm{y}}_{i}$ is subject to attraction $\mathcal{A}_{i}$ and repulsion $\mathcal{R}_{i}$ at each update, expressed as follows:
\begin{align}
    \bm{\mathrm{y}}_{i}\leftarrow\bm{\mathrm{y}}_{i}+\mathcal{A}_{i}+\mathcal{R}_{i},\label{eq:umap_update}
\end{align}
where the attraction $\mathcal{A}_{i}$ and the repulsion $\mathcal{R}_{i}$ are defined as
\begin{align}
    \mathcal{A}_{i}&=-\sum_{j\neq i}2ab||\bm{\mathrm{y}}_{i}-\bm{\mathrm{y}}_{j}||_{2}^{2(b-1)}v_{ij}w_{ij}(\bm{\mathrm{y}}_{i}-\bm{\mathrm{y}}_{j}),\label{eq:attraction}\\
    \mathcal{R}_{i}&=\sum_{j\neq i}\frac{2b}{||\bm{\mathrm{y}}_{i}-\bm{\mathrm{y}}_{j}||_{2}^{2}}v_{ij}(1-w_{ij})(\bm{\mathrm{y}}_{i}-\bm{\mathrm{y}}_{j}).\label{eq:repulsion}
\end{align}
The derivation of these forces is provided in Appendix~\ref{apx:att_rep}. In practice, the attraction and repulsion forces are approximated following the Monte-Carlo estimation of Eqs.~\eqref{eq:attraction} and~\eqref{eq:repulsion}. This attraction-repulsion framework provides a foundation for the UMAP algorithm and facilitates its modification. Notably, the global preservation capability in UMAP solely depends on the initialization.
\paragraph{Summary.} 
The UMAP algorithm is composed of the following three steps: 
\begin{enumerate}[label=\roman*., topsep=-0.0em, itemsep=-0.0em]
    \item \textbf{$k$NN graph construction} using Eqs.~\eqref{eq:wij} and~\eqref{eq:wji}.
    \item \textbf{Initialization} by Laplacian eigenmaps or PCA.
    \item \textbf{Optimization} following the update procedure in Eq.~\eqref{eq:umap_update}.
\end{enumerate}

\begin{figure*}
    \centering
    \includegraphics[width=165mm]{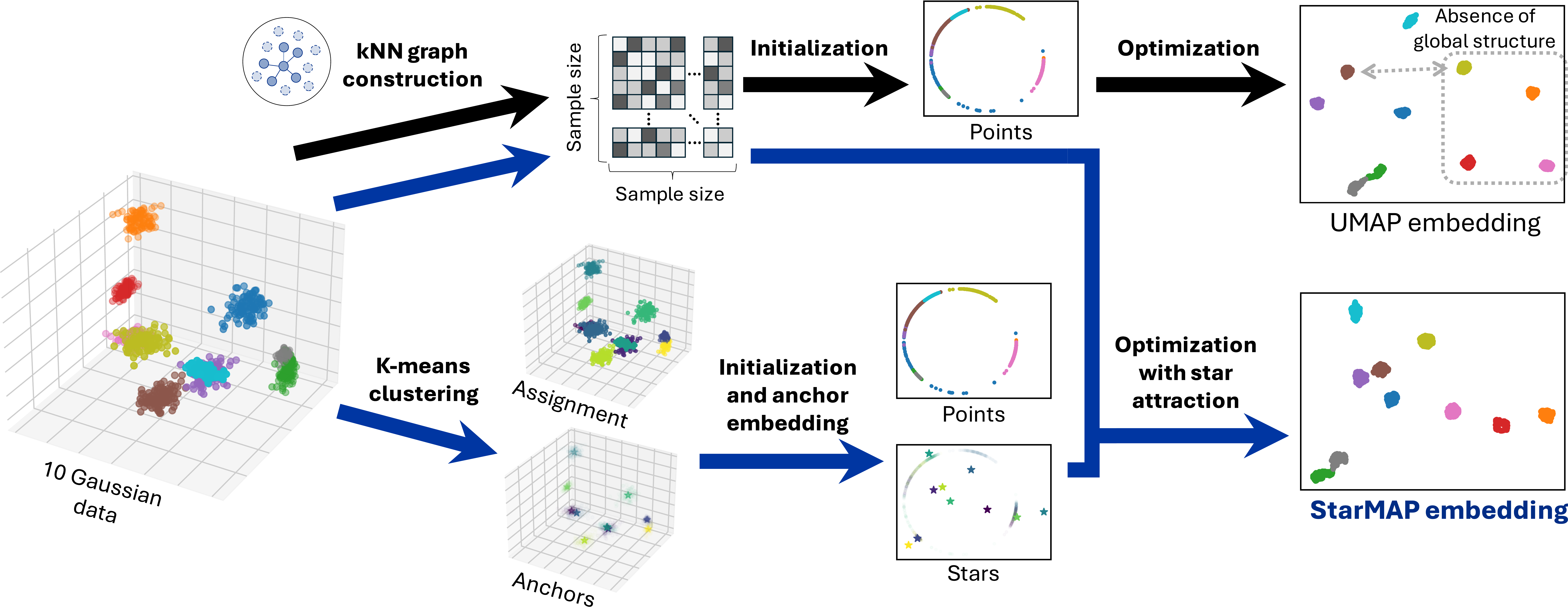}
    \caption{Illustration of the UMAP (\textbf{black}) and the proposed StarMAP (\textbf{\textcolor{myblue}{blue}}) algorithms. }
    \vspace{-0.3cm}
    \label{fig:pm}
\end{figure*}

\section{StarMAP}
\label{sec:starmap}
StarMAP begins by performing $K$-means clustering on the observed data to identify the anchor points and assign each point to the respective anchors. Then, StarMAP simultaneously embeds the observed data and the anchor points using PCA, where the embedding corresponding to the anchor points is referred to as stars. During the optimization process, StarMAP keeps the stars fixed and incorporates a star attraction in addition to the attraction and repulsion defined in Eqs.~\eqref{eq:attraction} and~\eqref{eq:repulsion}. StarMAP enhances the faithfulness of the resulting embedding by integrating the advantage of PCA. Figure~\ref{fig:pm} compares the UMAP and StarMAP methods.

\subsection{$K$-means Clustering}
StarMAP identifies the anchors and the assignment of each point from $K$-means clustering. These anchor points are denoted $\bm{\mathrm{A}}=[\bm{\mathrm{a}}_{1},\,\bm{\mathrm{a}}_{2},\,\ldots,\bm{\mathrm{a}}_{C}]^{\top}\in\mathbb{R}^{C \times D}$ where $C$ is the number of anchor points. The assignment function is difined as $m(\cdot)$ such that $m(i) = c$ if sample $i$ belongs to anchor $\bm{\mathrm{a}}_{c}\,(c=1,\,2,\,\ldots, C)$. Several methods for the automatic estimation of $C$ exist~\cite{pelleg2000x, schubert2023stop}; however these methods tend to be sensitive to noise and increase the computational complexity. Thus, StarMAP treats $C$ as a hyperparameter.

\subsection{Initialization and Anchor Embedding}
Following $K$-means clustering, StarMAP initializes $\bm{\mathrm{Y}}$ and embeds the anchor points using PCA. Here, StarMAP computes the dominant principal components of the matrix $[\bm{\mathrm{x}}_{1},\,\bm{\mathrm{x}}_{2},\,\ldots,\bm{\mathrm{x}}_{N},\bm{\mathrm{a}}_{1},\,\bm{\mathrm{a}}_{2},\ldots,\bm{\mathrm{a}}_{C}]^{\top}\in\mathbb{R}^{(N+C) \times D}$. Then, it assigns the first $N$ components as the initial embedding and the subsequent $C$ components as the stars $\bm{\mathrm{S}}=[\bm{\mathrm{s}}_{1},\,\bm{\mathrm{s}}_{2},\,\ldots,\bm{\mathrm{s}}_{C}]^{\top}\in\mathbb{R}^{C\times Q}$. The stars serve as an approximation of the PCA embedding, which enables StarMAP to preserve the global structure effectively.

\subsection{Optimization with Star Attraction}
Finally, StarMAP optimizes the embedding by incorporating the star attraction, which is computed as follows:
\begin{align}
    \mathcal{S}_{i}=-\frac{2ab||\bm{\mathrm{y}}_{i}-\bm{\mathrm{s}}_{m(i)}||_{2}^{2(b-1)}}{1+a||\bm{\mathrm{y}}_{i}-\bm{\mathrm{s}}_{m(i)}||_{2}^{2b}}d_{i}(\bm{\mathrm{y}}_{i}-\bm{\mathrm{s}}_{m(i)}),\label{eq:star_attraction}
\end{align}
where $d_{i}=\sum_{j \neq i}w_{ij}$ represents the degree of the node $i$. Note that the star attraction shares the same functional form as the neighbor attraction in Eq.~\eqref{eq:attraction}. StarMAP defines the star attraction in proportion to the degree; thus, it samples the star attraction concurrently with the neighbor attraction $\mathcal{A}_{i}$ in the implementation. Then, StarMAP introduces a new hyperparameter $\lambda\in[0,1]$ to balance the two forces, combining them as $\lambda\mathcal{S}_{i}+(1-\lambda)\mathcal{A}_{i}$. Thus, each update step of StarMAP is performed as follows:
\begin{wrapfigure}{r}{0.5\textwidth}
    \centering
    {
        \includegraphics[width=80mm]{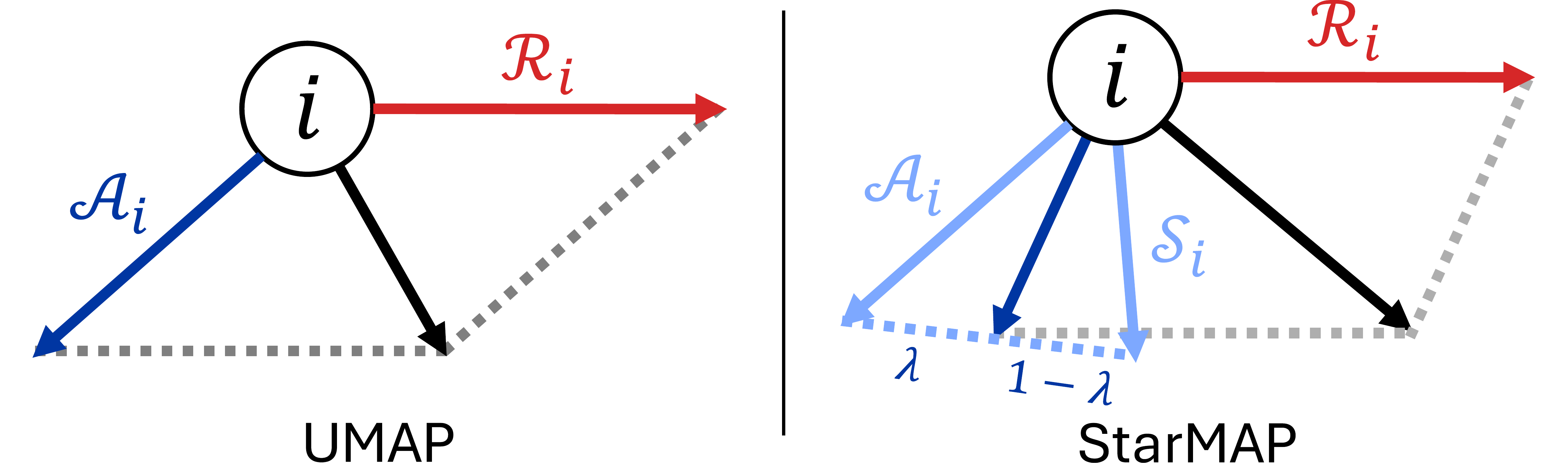}
    }
    \caption{Comparison of update procedure between UMAP and StarMAP.}
    \label{fig:comparison}
    \vspace{0.2cm}
\end{wrapfigure}
\begin{align}
    \bm{\mathrm{y}}_{i}\leftarrow\bm{\mathrm{y}}_{i}+\lambda\mathcal{S}_{i}+(1-\lambda)\mathcal{A}_{i}+\mathcal{R}_{i}.\label{eq:update_starmap}
\end{align}
The key distinction between StarMAP and UMAP lies in the computation of the attraction force (Figure~\ref{fig:comparison}). StarMAP incorporates the star attraction, which preserves the global information of the stars, whereas UMAP only considers the neighbor attraction on the $k$NN graph. This modification allows StarMAP to preserve both the global and local structures faithfully because it considers the global structure of the anchor points and the local structure of the neighborhood graph. In addition, StarMAP can be considered a neighbor embedding method that operates within the PCA subspace, providing a suboptimal solution to the interpretability issue associated with the UMAP embedding, as highlighted in the literature~\cite{mcinnes2018umap}.

\subsection{Time Complexity of StarMAP Algorithm}
We assess the time complexity of StarMAP, which is crucial when handling large-scale datasets. The time complexity of UMAP is $O(N^{1.14})$ to construct the approximated $k$NN graph~\citep{dong2011efficient}. The complexity of $K$-means clustering is linear with respect to $N$; thus, the complexity of the StarMAP algorithm is still $O(N^{1.14})$. However, the time complexity of $K$-means clustering increases linearly with the number of dimensions $D$. To address this issue, we perform PCA before $K$-means clustering when $\bm{\mathrm{Y}}$ has a large number of dimensions (e.g., $D\geq300$).
\paragraph{Summary.} 
The StarMAP algorithm is composed of the following steps: 
\begin{enumerate}[label=\roman*., topsep=-0.0em, itemsep=-0.0em]
    \item \textbf{$k$NN graph construction} using Eqs.~\eqref{eq:wij} and~\eqref{eq:wji}.
    \item \textbf{$K$-means clustering} to determine the anchor points and the assignment function.
    \item \textbf{Initialization and anchor embedding} via PCA.
    \item \textbf{Optimization with star attraction} following the update procedure in Eq.~\eqref{eq:update_starmap}.
\end{enumerate}
\section{Experiments}
\label{sec:experiment}
\subsection{Visualization of Synthetic Data}
\begin{figure*}
    \centering
    \includegraphics[width=165mm]{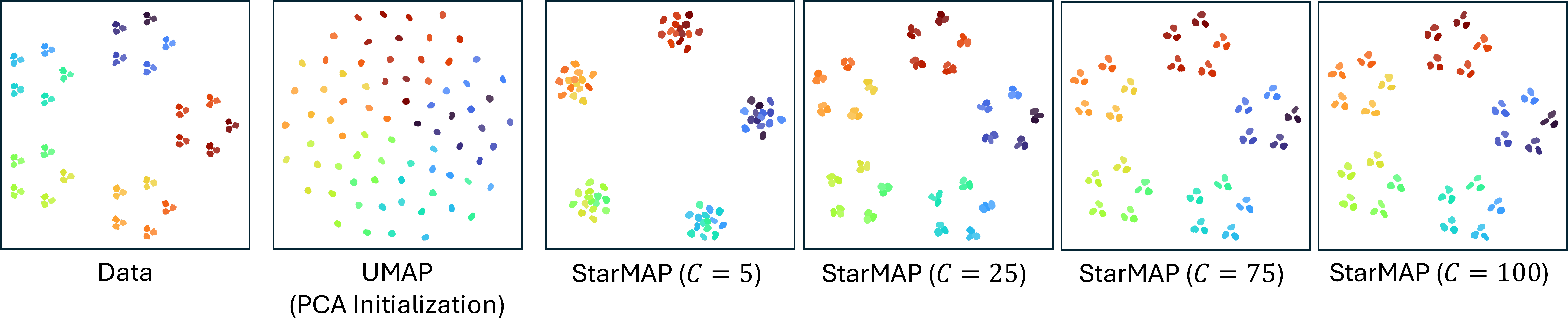}
    \vspace{-0.2cm}
    \caption{Visualization result obtained by UMAP and StarMAP on synthetic hierarchical cluster dataset.}
    \label{fig:exp_synthesis}
\end{figure*}

A synthetic hierarchical cluster dataset was created to evaluate our method. This dataset contains five large clusters, each of which is subdivided into five intermediate clusters, which are divided into three small clusters (Figure~\ref{fig:exp_synthesis}, \textit{left}). The smallest clusters comprise $100$ data points, resulting in a total of $N=7,500$ points. The number of dimensions is $2$ ($D=2$), providing a known ground truth for DR. StarMAP was compared with UMAP (PCA initialization). The StarMAP algorithm introduces two novel hyperparameters, i.e., the number of anchor points $C$ and the proportion $\lambda$ in Eq.~\eqref{eq:update_starmap} to combine the attraction forces. The number of the smallest clusters is $75$; thus, $C=75$ was a reasonable choice. However, the optimal choice of $C$ is not immediately obvious in the general unsupervised DR context. Therefore, we tested its effect using different values, i.e., $C=5,\,25,\,75,\,100$. We found that the appropriate value of $\lambda$ was invariant across the datasets, similar to most UMAP parameters; thus $\lambda=0.1$ was fixed in this study.
\par
Figure~\ref{fig:exp_synthesis} shows the UMAP and StarMAP visualizations. The result obtained by UMAP highlights a potential limitation of neighbor embedding because it failed to preserve the global similarities between the clusters. Despite using the PCA initialization, which was applied in a previous study~\cite{kobak2021initialization}, UMAP failed to capture the structures of the intermediate and the largest clusters, resulting in a misleading visualization. In contrast, StarMAP effectively preserved the global similarities. StarMAP accurately identified the largest clusters with $C=5$ and the intermediate clusters with $C=25$. In addition, the smallest clusters were recognized with $C=75$ and $C=100$, highlighting the robustness of StarMAP when $C$ is set larger than the optimal value ($C=75$). However, StarMAP produced different visualizations for $C=5,\,25,\,75$. Thus, tuning $C$ is essential when applying StarMAP to the general unsupervised DR tasks. 
\subsection{Visualization of Real-World Data}
\label{subsec:real_world_data_vis}
The following six datasets were used to validate different DR methods:
\begin{enumerate}[label=\Roman*., topsep=-0.0em, itemsep=-0.0em]
   \item\textbf{Mammoth}~\cite{wang2021understanding}.
   This dataset consists of 3D points of the mammoth skeleton ($N=20,000,\,D=3$). A previous study attempted to preserve the interbody part similarities~\cite{wang2021understanding}. 
   \item\textbf{MNIST}.
    This dataset contains $28 \times 28$ grayscale images of handwritten digits and corresponding labels ($N=60,000,\,D=784$). Previous studies focused on identifying the digit clusters, e.g.,~\cite{mcinnes2018umap}.
   \item\textbf{Fashion MNIST}~\cite{xiao2017fashion}.
    This dataset contains $28 \times 28$ grayscale images of fashion items and labels indicating the specific classes of each item ($N=60,000,\,D=784$). Previous research has focused on identifying the item clusters; however, Fashion MNIST is more complicated than MNIST.
    \item\textbf{Retina}~\cite{macosko2015highly}.
    This dataset consists of mouse retina scRNA-seq data provided by Macosko et al.~\cite{macosko2015highly} ($N=44,808,\, D=400$). It includes multiple populations of retinal cells, e.g., photoreceptor and amacrine cells. Here, each cell cluster must be identified~\cite{kobak2019art}.
   \item\textbf{Neocortex}~\cite{tasic2018shared}.
    This dataset contains scRNA-seq data of adult mouse neocortex cells provided by Tasic et al.~\cite{tasic2018shared} ($N=23,822,\, D=3,000$), and it features a hierarchical cluster structure comprising inhibitory, excitatory, and non-neuronal cell populations. Previous research has attempted to identify the cell clusters while preserving the similarities between the large groups~\cite{kobak2019art}.
   \item\textbf{Planaria}~\citep{plass2018cell}.
This dataset contains scRNA-seq data exhibiting the lineage during cell differentiation from intermediate cellular states to terminally differentiated cell types ($N=21,612,\, D=50$)~\cite{plass2018cell}. Previous research has aimed to preserve the lineage while identifying the terminal cell clusters~\cite{klimovskaia2020poincare}.
\end{enumerate}
In this evaluation, we compared StarMAP with PCA~\citep{hotelling1933analysis}, t-SNE~\citep{van2008visualizing, van2014accelerating}, and UMAP~\citep{mcinnes2018umap} as baselines, as well as with PHATE~\citep{moon2019visualizing} and PaCMAP~\citep{wang2021understanding} which are previous methods designed to preserve the global structure. Default settings were used for each DR method. We fixed $\lambda=0.1,\,k=20$ across all datasets for StarMAP, and the number of anchor points was set to $C=60,\,12,\,15,\,150,\,100,\,200$ for the Mammoth, MNIST, Fashion MNIST, Retina, Cortex, and Planaria datasets, respectively. We compared the two-dimensional embedding and averaged metric scores over 10 runs. Here, we evaluated the methods using two metrics, i.e., the $k$NN classification accuracy (\textit{local}) and the distance correlation (\textit{global}). The metric selection was based on the computational feasibility and available trustworthy information (e.g., class labels). The $k$NN accuracy was calculated using the two-dimensional embedding of each method with $k=5$. Note that, the Shepard goodness~\citep{joia2011local}, which is a global metric analogous to the distance correlation, while widely used, is computationally intractable for large-scale datasets due to its $O(N^{2})$ space complexity for rank computation of the distance matrices. Thus, we employed the correlation coefficient between the distance matrices to facilitate a scalable and effective global evaluation.
\begin{figure*}[t]
    \centering
    \includegraphics[width=165mm]{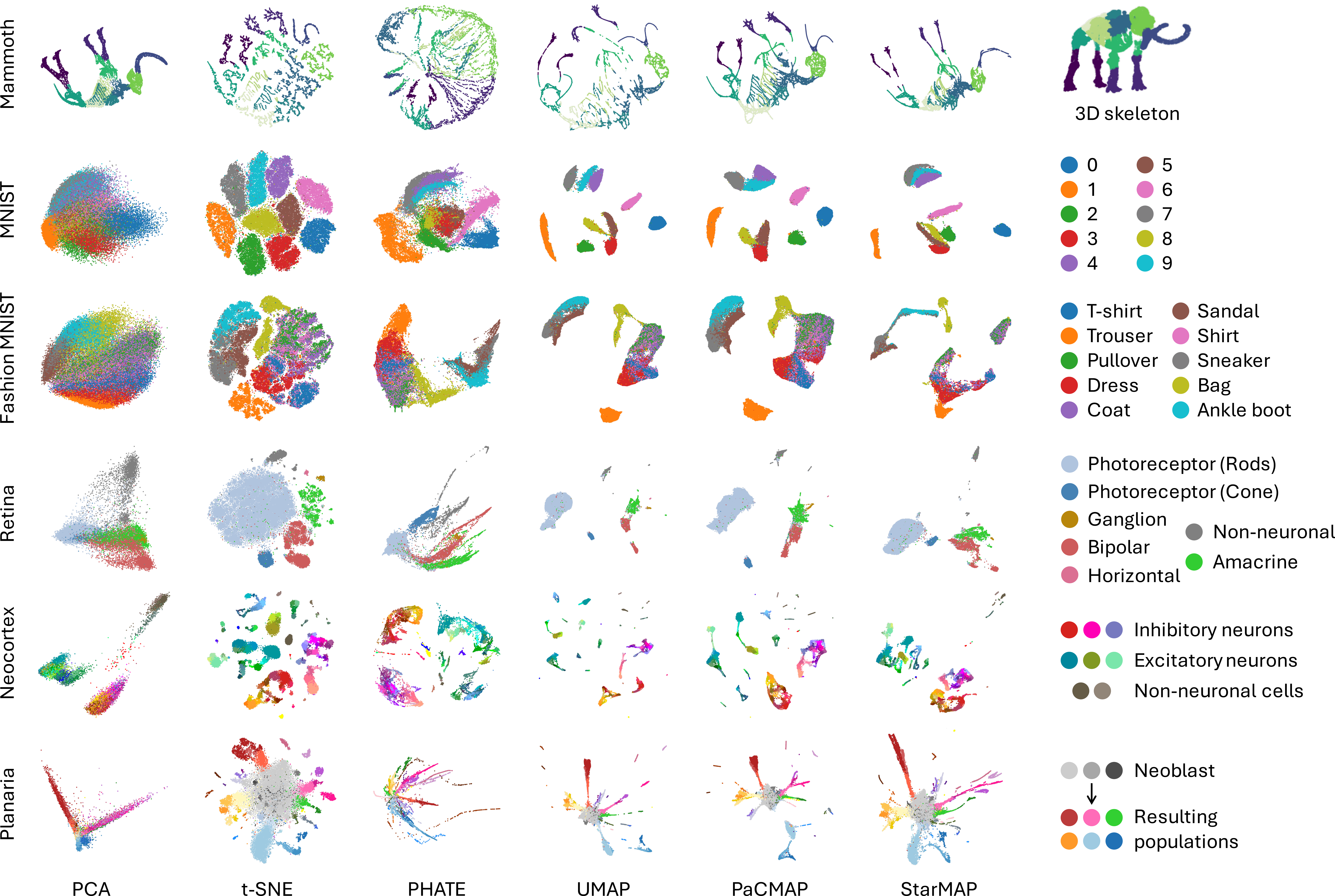}
    \caption{Visualization results on six real-world datasets. The color code of the Mammoth dataset reflects different body parts. For the MNIST, Fashion MNIST, and scRNA-seq datasets, the codes identify the digit, fashion item, and cel clusters, respectively. In addition, the code of the Cortex dataset exhibits the hierarchical structure within the inhibitory (hot), excitatory (cool), and non-neuron (gray) clusters, and that of the Planaria dataset contains the lineage from the neoblast (gray) to the resulting populations.}
    \label{fig:exp_visualization}
\end{figure*}
\par
Figure~\ref{fig:exp_visualization} presents the visualization results obtained on all datasets. On the Mammoth dataset, both PCA and StarMAP successfully preserved the shape, whereas, on the MNIST dataset, the neighbor embedding methods and StarMAP identified the digit clusters effectively. Thus, only StarMAP provided faithful visualizations for both datasets, demonstrating its advantage over the compared DR methods. On the Fashion MNIST dataset, StarMAP effectively separated the item clusters, achieving competitive results with the neighbor embedding methods. In addition, all compared methods identified the large clusters in the Retina dataset successfully; however, the PCA, PHATE, and PaCMAP embedding exhibited overlap between the horizontal and amacrine cell populations. In contrast, t-SNE, UMAP, and StarMAP distinguished these populations. t-SNE produced the most interpretable embedding for this dataset, as the ganglion and horizontal clusters were more spread out. This result aligns with prior studies that indicated t-SNE's low attraction force compared to UMAP~\cite{bohm2022attraction, jenssen2024map}. 
On the Neocortex dataset, PCA preserved the similarities between the inhibitory, excitatory, and non-neuronal clusters; however, it failed to separate the clusters within these large groups. Conversely, t-SNE, PHATE, UMAP, and PaCMAP resolved the small cell clusters successfully but did not retain the similarities within the large groups. In contrast, StarMAP preserved both the small cell clusters and the global similarities among the large groups, thereby demonstrating its effectiveness in terms of maintaining the global and local structures. On the Planaria dataset, t-SNE, UMAP, and PaCMAP identified some terminal populations but failed to capture the lineage during the differentiation process. Here, PCA, PHATE, and StarMAP exhibited competitive performance in preserving the lineage structure. Notably, StarMAP placed the neoblast populations at the center of the embedding and distinctly separated each terminal population, highlighting its ability to balance lineage preservation with clear cluster separation.
\begin{figure*}
    \centering
    \includegraphics[width=165mm]{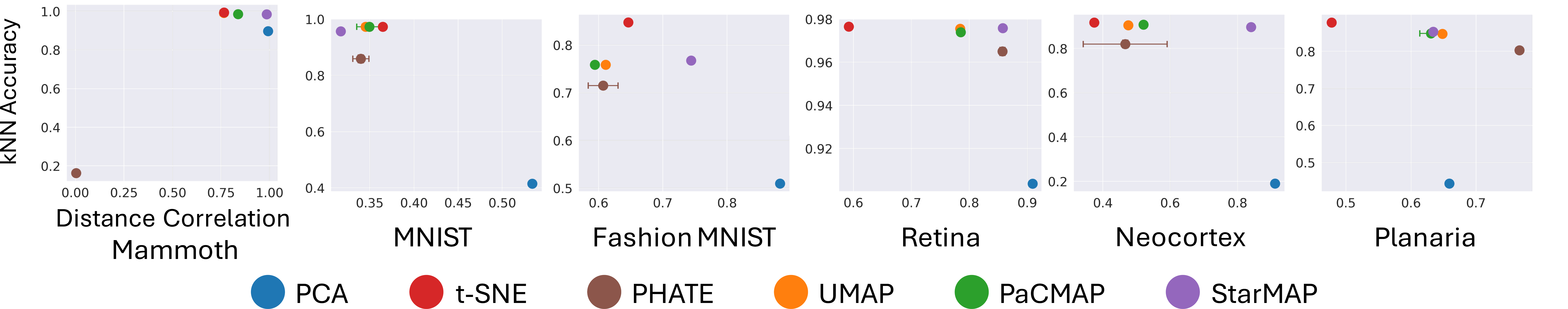}
    \vspace{-0.2cm}
    \caption{Quantitative results of error bar plots with distance correlation (\textit{global}, \textit{x-axis}) and $k$NN accuracy (\textit{local}, \textit{y-axis}).}
    \label{fig:exp_error_bar}
\end{figure*}
\par
Figure~\ref{fig:exp_error_bar} shows the error bar plots for all datasets. Overall, all PCA embeddings achieved high distance correlation scores, while the neighbor embedding methods achieved high $k$NN accuracy. This observation supports our motivation to integrate the advantage of PCA into the neighbor embedding techniques. StarMAP consistently obtained relatively high scores across both metrics on the Mammoth, Fashion MNIST, and scRNA-seq datasets with remarkable stability. Notably, StarMAP appeared in the upper-right corner of the error bar plots for the Mammoth, Fashion MNIST, Retina, and Neocortex datasets, highlighting the faithfulness of the embeddings. On the Planaria dataset, PHATE, UMAP, PaCMAP, and StarMAP obtained competitive scores; however, StarMAP demonstrated a unique advantage by effectively capturing the cell lineage during the differentiation process, thereby underscoring its utility. On the MNIST dataset, StarMAP did not outperform the compared neighbor embedding methods since StarMAP depends on the first two principal components which overlapped some digit clusters. The first two components were insufficient in some cases to fully represent the data structure, which resulted in degraded accuracy on MNIST. However, the distance correlation scores of the neighbor embedding methods were competitive, and StarMAP's visualization for MNIST was comparable to those of the compared neighbor embedding methods (apart from the overlapping issue). In addition, Appendix~\ref{apx:mnist} demonstrates that StarMAP, when configured with different $C$ values, can achieve better distance correlation scores than the previous neighbor embedding methods. Holistically, StarMAP is superior or at least competitive in both visualization and quantitative evaluations. Visualizations of the star position are provided in Appendix~\ref{apx:star_vis}, and enlarged results with additional comparisons are presented in Appendix~\ref{apx:vis_comparison}.
\subsection{Visualization of Deep Representation}
Finally, we visualized the representation of a deep neural network model using StarMAP. We selected CLIP~\citep{radford2021learning, sun2023eva}, which is a commonly used model with image and text encoders. These encoders share a latent space, and CLIP is trained to align features through image-text associations. The CLIP representation has exhibited strong zero-shot capabilities, thereby making it suitable for various downstream tasks~\cite{ramesh2022hierarchical}. In addition, its exploration has garnered significant attention~\cite {shen2022how}, with prevailing findings indicating that CLIP can effectively extract detailed semantics from images~\cite{zhou2022extract} and produces \textit{semantically coherent} features~\cite{sun2024alpha}. In this study, we confirmed this by visualizing the image representation. The CIFAR100~\citep{krizhevsky2009learning} dataset, which contains $50,000$ images, $100$ fine labels (e.g., ``bicycle” and ``bee”), and $20$ super labels (e.g., ``vehicles 1” and ``insects”), was used for this analysis. To illustrate the semantic coherence of the representation, we colored the image embedding based on its similarity to specific text features. This process was performed as follows.
\begin{enumerate}[label=\roman*., topsep=-0.0em, itemsep=-0.0em]
    \item The image features of CIFAR100 and the text features of a certain text (e.g., ``\texttt{a photo of vehicles}”) were extracted using the CLIP encoders.
    \item The similarity scores between the image and text features were computed.
    \item The image features were embedded into the two-dimensional space and colored with the similarity scores.
\end{enumerate}
If the embedding preserves the CLIP representation faithfully, it will naturally maintain the semantic coherence in the embedding. To test this, we selected three text queries: ``\texttt{a photo of vehicles}," ``\texttt{a photo of aquatic mammals}," and ``\texttt{a photo of plants}," which correspond to the super labels of the CIFAR100 dataset. We selected UMAP with the PCA initialization and StarMAP with $\lambda=0.1$ and $C=300$. Then, the embeddings were evaluated in terms of their ability to preserve semantic coherence, which was reflected in how well the image embeddings aligned with the image-text similarities.
\begin{figure*}
    \centering
    \includegraphics[width=165mm]{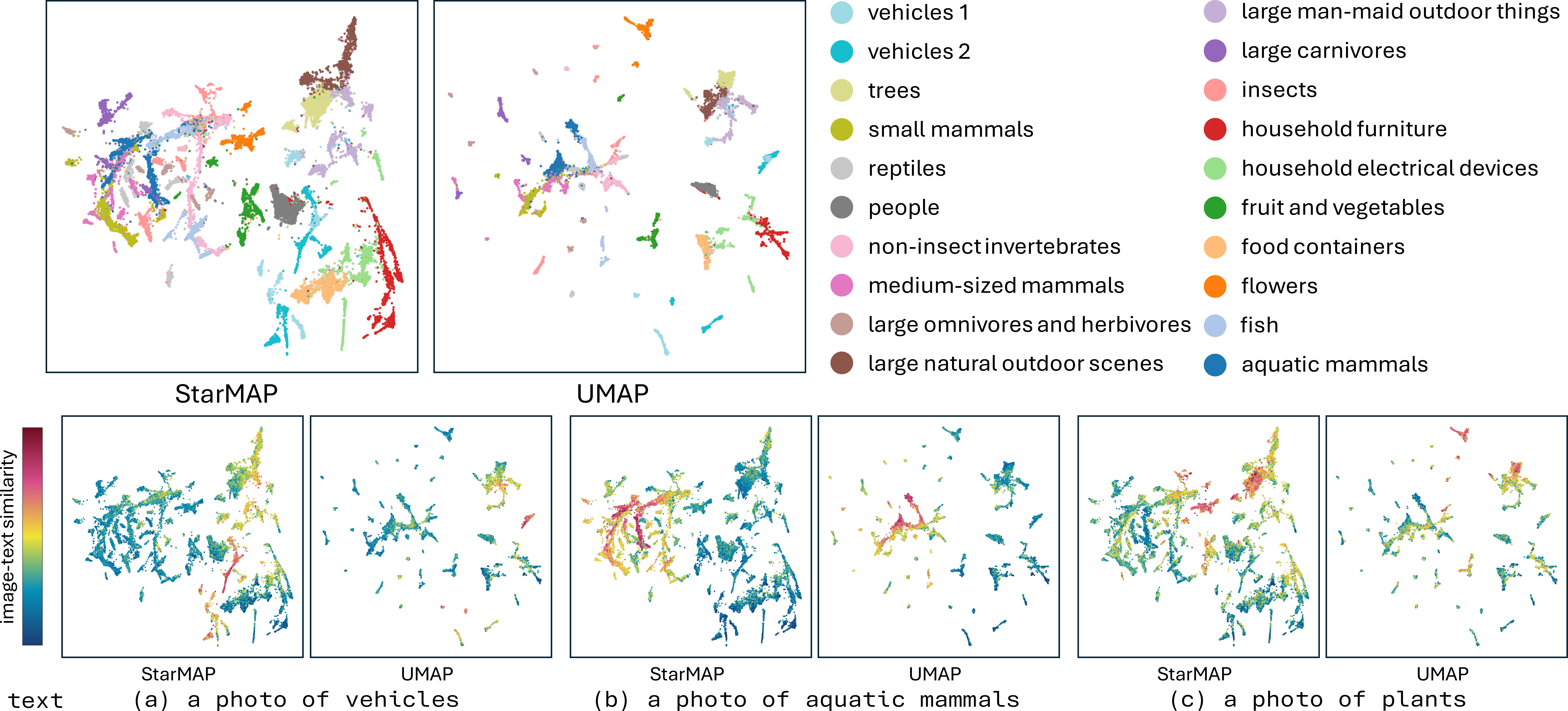}
    \caption{Visualization results in CIFAR100 features extracted by CLIP. We colored the embeddings along the super labels of CIFAR100 (\textit{upper}) and the similarity scores between the image and text features (\textit{bottom}).}
    \label{fig:exp_clip_vis}
\end{figure*}
\par
Figure~\ref{fig:exp_clip_vis} shows the visualization results of the CLIP representation. The embeddings colored by the super labels reveal that StarMAP produced a more compact visualization than UMAP. This difference influenced the overall visualization when using the similarity scores. When visualizing with the text ``\texttt{a photo of vehicles}," the StarMAP embedding exhibited a clear vertical alignment, and this alignment grouped categories, e.g., ``vehicles 1" (including ``bicycle" and ``bus"), ``vehicles 2" (including ``tractor" and ``tank"), and ``large man-made outdoor things" (including ``road" and ``bridge"), which was not as clearly evident in the UMAP embedding. In the visualization of ``\texttt{a photo of aquatic mammals}," StarMAP highlighted the gradation within the animal-related classes, effectively aligning the embedding with high similarity. For the ``\texttt{a photo of plants}" visualization, StarMAP maintained smooth preservation of the semantic similarities between the plant-related classes. Overall, StarMAP outperformed UMAP in preserving the semantic coherence of the CLIP representation, thereby demonstrating better alignment of the image-text similarities. 

\section{Conclusion}
This paper has proposed StarMAP, which is designed to realize faithful visualizations of high-dimensional data. StarMAP introduces the concept of stars and their associated attraction forces, which play a key role in preserving the global structure of the data.
\par
The experimental results demonstrate its effectiveness but also exhibit some limitations. (\rom{1}) StarMAP requires an additional hyperparameter $C$ that controls the number of stars whose selection impacts the visualization; thus, a clear criterion for selecting this parameter is required. A heuristic solution to this issue is given in Appendix~\ref{apx:heuristic}. (\rom{2}) The StarMAP embedding is slightly less effective when the PCA results in overlapping clusters. Therefore, a dedicated algorithm is required to visualize such data.
\par
Recent deep models have demonstrated high accuracy across various downstream tasks; thus, understanding their representations is crucial for developing algorithms with sufficient explainability. Although t-SNE and UMAP are commonly used to visualize internal representation, the experiment results obtained by these methods fail to preserve global similarities, frequently resulting in misleading visualizations. We believe that StarMAP will contribute to understanding large-scale data structure and deep representations, thereby offering valuable insights for explainable AI. 

\bibliography{paper}

\begin{thebibliography}{45}
\providecommand{\natexlab}[1]{#1}
\providecommand{\url}[1]{\texttt{#1}}
\expandafter\ifx\csname urlstyle\endcsname\relax
  \providecommand{\doi}[1]{doi: #1}\else
  \providecommand{\doi}{doi: \begingroup \urlstyle{rm}\Url}\fi

\bibitem[Amid \& Warmuth(2019)Amid and Warmuth]{amid2019trimap}
Amid, E. and Warmuth, M.~K.
\newblock {TriMap}: Large-scale dimensionality reduction using triplets.
\newblock \emph{arXiv preprint arXiv:1910.00204}, 2019.

\bibitem[Artemenkov \& Panov(2020)Artemenkov and Panov]{artemenkov2020ncvis}
Artemenkov, A. and Panov, M.
\newblock Ncvis: noise contrastive approach for scalable visualization.
\newblock In \emph{Proceedings of the International Conference on World Wide Web}, pp.\  2941--2947, 2020.

\bibitem[Balasubramanian \& Schwartz(2002)Balasubramanian and Schwartz]{balasubramanian2002isomap}
Balasubramanian, M. and Schwartz, E.~L.
\newblock The isomap algorithm and topological stability.
\newblock \emph{Science}, 295\penalty0 (5552):\penalty0 7--7, 2002.

\bibitem[Belkin \& Niyogi(2003)Belkin and Niyogi]{belkin2003laplacian}
Belkin, M. and Niyogi, P.
\newblock Laplacian eigenmaps for dimensionality reduction and data representation.
\newblock \emph{Neural Computation}, 15\penalty0 (6):\penalty0 1373--1396, 2003.

\bibitem[B{\"o}hm et~al.(2022)B{\"o}hm, Berens, and Kobak]{bohm2022attraction}
B{\"o}hm, J.~N., Berens, P., and Kobak, D.
\newblock Attraction-repulsion spectrum in neighbor embeddings.
\newblock \emph{Journal of Machine Learning Research}, 23\penalty0 (95):\penalty0 1--32, 2022.

\bibitem[B{\"o}hm et~al.(2023)B{\"o}hm, Berens, and Kobak]{bohm2023unsupervised}
B{\"o}hm, N., Berens, P., and Kobak, D.
\newblock Unsupervised visualization of image datasets using contrastive learning.
\newblock In \emph{Proceedings of the International Conference on Learning Representations}, pp.\  1--21, 2023.

\bibitem[Coifman \& Lafon(2006)Coifman and Lafon]{coifman2006diffusion}
Coifman, R.~R. and Lafon, S.
\newblock Diffusion maps.
\newblock \emph{Applied and Computational Harmonic Analysis}, 21\penalty0 (1):\penalty0 5--30, 2006.

\bibitem[Damrich et~al.(2022)Damrich, B{\"o}hm, Hamprecht, and Kobak]{damrich2022t}
Damrich, S., B{\"o}hm, J.~N., Hamprecht, F.~A., and Kobak, D.
\newblock From {$ t $-SNE} to {UMAP} with contrastive learning.
\newblock \emph{arXiv preprint arXiv:2206.01816}, 2022.

\bibitem[Dong et~al.(2011)Dong, Moses, and Li]{dong2011efficient}
Dong, W., Moses, C., and Li, K.
\newblock Efficient k-nearest neighbor graph construction for generic similarity measures.
\newblock In \emph{Proceedings of the international conference on World Wide Web}, pp.\  577--586, 2011.

\bibitem[Fu et~al.(2019)Fu, Zhang, Cai, and Ren]{fu2019atsne}
Fu, C., Zhang, Y., Cai, D., and Ren, X.
\newblock {AtSNE}: Efficient and robust visualization on {GPU} through hierarchical optimization.
\newblock In \emph{Proceedings of the ACM SIGKDD International Conference on Knowledge Discovery \& Data Mining}, pp.\  176--186, 2019.

\bibitem[Hotelling(1933)]{hotelling1933analysis}
Hotelling, H.
\newblock Analysis of a complex of statistical variables into principal components.
\newblock \emph{Journal of Educational Psychology}, 24\penalty0 (6):\penalty0 417--441, 1933.

\bibitem[Huang et~al.(2024)Huang, Wang, and Rudin]{huang2024navigating}
Huang, H., Wang, Y., and Rudin, C.
\newblock Navigating the effect of parametrization for dimensionality reduction.
\newblock In \emph{Advances in Neural Information Processing Systems}, pp.\  1--43, 2024.

\bibitem[Jenssen(2024)]{jenssen2024map}
Jenssen, R.
\newblock {MAP} {IT} to visualize representations.
\newblock In \emph{Proceedings of the International Conference on Learning Representations}, pp.\  1--35, 2024.

\bibitem[Joia et~al.(2011)Joia, Coimbra, Cuminato, Paulovich, and Nonato]{joia2011local}
Joia, P., Coimbra, D., Cuminato, J.~A., Paulovich, F.~V., and Nonato, L.~G.
\newblock Local affine multidimensional projection.
\newblock \emph{IEEE Transactions on Visualization and Computer Graphics}, 17\penalty0 (12):\penalty0 2563--2571, 2011.

\bibitem[Klimovskaia et~al.(2020)Klimovskaia, Lopez-Paz, Bottou, and Nickel]{klimovskaia2020poincare}
Klimovskaia, A., Lopez-Paz, D., Bottou, L., and Nickel, M.
\newblock Poincar{\'e} maps for analyzing complex hierarchies in single-cell data.
\newblock \emph{Nature Communications}, 11\penalty0 (1):\penalty0 2966, 2020.

\bibitem[Kobak \& Berens(2019)Kobak and Berens]{kobak2019art}
Kobak, D. and Berens, P.
\newblock The art of using {t-SNE} for single-cell transcriptomics.
\newblock \emph{Nature Communications}, 10\penalty0 (1):\penalty0 5416, 2019.

\bibitem[Kobak \& Linderman(2021)Kobak and Linderman]{kobak2021initialization}
Kobak, D. and Linderman, G.~C.
\newblock Initialization is critical for preserving global data structure in both {t-SNE} and {UMAP}.
\newblock \emph{Nature Biotechnology}, 39\penalty0 (2):\penalty0 156--157, 2021.

\bibitem[Krizhevsky(2009)]{krizhevsky2009learning}
Krizhevsky, A.
\newblock Learning multiple layers of features from tiny images.
\newblock Technical report, University of Toronto, 2009.

\bibitem[Li et~al.(2024)Li, Wang, Chen, Shen, and Chao]{li2024fedne}
Li, Z., Wang, X., Chen, H.-Y., Shen, H.~W., and Chao, W.-L.
\newblock {FedNE}: Surrogate-assisted federated neighbor embedding for dimensionality reduction.
\newblock In \emph{Advances in Neural Information Processing Systems}, pp.\  1--27, 2024.

\bibitem[Macosko et~al.(2015)Macosko, Basu, Satija, Nemesh, Shekhar, Goldman, Tirosh, Bialas, Kamitaki, Martersteck, et~al.]{macosko2015highly}
Macosko, E.~Z., Basu, A., Satija, R., Nemesh, J., Shekhar, K., Goldman, M., Tirosh, I., Bialas, A.~R., Kamitaki, N., Martersteck, E.~M., et~al.
\newblock Highly parallel genome-wide expression profiling of individual cells using nanoliter droplets.
\newblock \emph{Cell}, 161\penalty0 (5):\penalty0 1202--1214, 2015.

\bibitem[May(1992)]{may1992simplicial}
May, J.~P.
\newblock \emph{Simplicial objects in algebraic topology}, volume~11.
\newblock University of Chicago Press, 1992.

\bibitem[McInnes et~al.(2018)McInnes, Healy, and Melville]{mcinnes2018umap}
McInnes, L., Healy, J., and Melville, J.
\newblock {UMAP}: Uniform manifold approximation and projection for dimension reduction.
\newblock \emph{arXiv preprint arXiv:1802.03426}, 2018.

\bibitem[Moon et~al.(2019)Moon, Van~Dijk, Wang, Gigante, Burkhardt, Chen, Yim, Elzen, Hirn, Coifman, et~al.]{moon2019visualizing}
Moon, K.~R., Van~Dijk, D., Wang, Z., Gigante, S., Burkhardt, D.~B., Chen, W.~S., Yim, K., Elzen, A. v.~d., Hirn, M.~J., Coifman, R.~R., et~al.
\newblock Visualizing structure and transitions in high-dimensional biological data.
\newblock \emph{Nature Biotechnology}, 37\penalty0 (12):\penalty0 1482--1492, 2019.

\bibitem[Pedregosa et~al.(2011)Pedregosa, Varoquaux, Gramfort, Michel, Thirion, Grisel, Blondel, Prettenhofer, Weiss, Dubourg, et~al.]{pedregosa2011scikit}
Pedregosa, F., Varoquaux, G., Gramfort, A., Michel, V., Thirion, B., Grisel, O., Blondel, M., Prettenhofer, P., Weiss, R., Dubourg, V., et~al.
\newblock Scikit-learn: Machine learning in {Python}.
\newblock \emph{Journal of Machine Learning Research}, 12:\penalty0 2825--2830, 2011.

\bibitem[Pelleg(2000)]{pelleg2000x}
Pelleg, D.
\newblock {X}-means: Extending {K}-means with efficient estimation of the number of clusters.
\newblock In \emph{Proceedings of the International Conference on Machine Learning}, pp.\  727--734, 2000.

\bibitem[Plass et~al.(2018)Plass, Solana, Wolf, Ayoub, Misios, Gla{\v{z}}ar, Obermayer, Theis, Kocks, and Rajewsky]{plass2018cell}
Plass, M., Solana, J., Wolf, F.~A., Ayoub, S., Misios, A., Gla{\v{z}}ar, P., Obermayer, B., Theis, F.~J., Kocks, C., and Rajewsky, N.
\newblock Cell type atlas and lineage tree of a whole complex animal by single-cell transcriptomics.
\newblock \emph{Science}, 360\penalty0 (6391):\penalty0 eaaq1723, 2018.

\bibitem[Radford et~al.(2021)Radford, Kim, Hallacy, Ramesh, Goh, Agarwal, Sastry, Askell, Mishkin, Clark, et~al.]{radford2021learning}
Radford, A., Kim, J.~W., Hallacy, C., Ramesh, A., Goh, G., Agarwal, S., Sastry, G., Askell, A., Mishkin, P., Clark, J., et~al.
\newblock Learning transferable visual models from natural language supervision.
\newblock In \emph{Proceedings of the International Conference on Machine Learning}, pp.\  8748--8763, 2021.

\bibitem[Ramesh et~al.(2022)Ramesh, Dhariwal, Nichol, Chu, and Chen]{ramesh2022hierarchical}
Ramesh, A., Dhariwal, P., Nichol, A., Chu, C., and Chen, M.
\newblock Hierarchical text-conditional image generation with {CLIP} latents.
\newblock \emph{arXiv preprint arXiv:2204.06125}, 2022.

\bibitem[Rudin et~al.(2022)Rudin, Chen, Chen, Huang, Semenova, and Zhong]{rudin2022interpretable}
Rudin, C., Chen, C., Chen, Z., Huang, H., Semenova, L., and Zhong, C.
\newblock Interpretable machine learning: Fundamental principles and 10 grand challenges.
\newblock \emph{Statistic Surveys}, 16:\penalty0 1--85, 2022.

\bibitem[Schubert(2023)]{schubert2023stop}
Schubert, E.
\newblock Stop using the elbow criterion for {K-means} and how to choose the number of clusters instead.
\newblock \emph{ACM SIGKDD Explorations Newsletter}, 25\penalty0 (1):\penalty0 36--42, 2023.

\bibitem[Shen et~al.(2022)Shen, Li, Tan, Bansal, Rohrbach, Chang, Yao, and Keutzer]{shen2022how}
Shen, S., Li, L.~H., Tan, H., Bansal, M., Rohrbach, A., Chang, K.-W., Yao, Z., and Keutzer, K.
\newblock How much can {CLIP} benefit vision-and-language tasks?
\newblock In \emph{Proceedings of the International Conference on Learning Representations}, pp.\  1--18, 2022.

\bibitem[Sun et~al.(2023)Sun, Fang, Wu, Wang, and Cao]{sun2023eva}
Sun, Q., Fang, Y., Wu, L., Wang, X., and Cao, Y.
\newblock {EVA-CLIP}: Improved training techniques for {CLIP} at scale.
\newblock \emph{arXiv preprint arXiv:2303.15389}, 2023.

\bibitem[Sun et~al.(2024)Sun, Fang, Wu, Zhang, Zang, Kong, Xiong, Lin, and Wang]{sun2024alpha}
Sun, Z., Fang, Y., Wu, T., Zhang, P., Zang, Y., Kong, S., Xiong, Y., Lin, D., and Wang, J.
\newblock Alpha-{CLIP}: A clip model focusing on wherever you want.
\newblock In \emph{Proceedings of the IEEE/CVF Conference on Computer Vision and Pattern Recognition}, pp.\  13019--13029, 2024.

\bibitem[Tang et~al.(2016)Tang, Liu, Zhang, and Mei]{tang2016visualizing}
Tang, J., Liu, J., Zhang, M., and Mei, Q.
\newblock Visualizing large-scale and high-dimensional data.
\newblock In \emph{Proceedings of the International Conference on World Wide Web}, pp.\  287--297, 2016.

\bibitem[Tasic et~al.(2018)Tasic, Yao, Graybuck, Smith, Nguyen, Bertagnolli, Goldy, Garren, Economo, Viswanathan, et~al.]{tasic2018shared}
Tasic, B., Yao, Z., Graybuck, L.~T., Smith, K.~A., Nguyen, T.~N., Bertagnolli, D., Goldy, J., Garren, E., Economo, M.~N., Viswanathan, S., et~al.
\newblock Shared and distinct transcriptomic cell types across neocortical areas.
\newblock \emph{Nature}, 563\penalty0 (7729):\penalty0 72--78, 2018.

\bibitem[Torgerson(1958)]{torgerson1958theory}
Torgerson, W.~S.
\newblock \emph{Theory and methods of scaling}.
\newblock Wiley, 1958.

\bibitem[Van Der~Maaten(2009)]{van2009learning}
Van Der~Maaten, L.
\newblock Learning a parametric embedding by preserving local structure.
\newblock In \emph{Proceedings of the International Conference on Artificial Intelligence and Statistics}, pp.\  384--391, 2009.

\bibitem[Van Der~Maaten(2014)]{van2014accelerating}
Van Der~Maaten, L.
\newblock Accelerating {t-SNE} using tree-based algorithms.
\newblock \emph{Journal of Machine Learning Research}, 15\penalty0 (1):\penalty0 3221--3245, 2014.

\bibitem[Van~der Maaten \& Hinton(2008)Van~der Maaten and Hinton]{van2008visualizing}
Van~der Maaten, L. and Hinton, G.
\newblock Visualizing data using {t-SNE}.
\newblock \emph{Journal of Machine Learning Research}, 9\penalty0 (11):\penalty0 1--25, 2008.

\bibitem[Van Der~Maaten \& Weinberger(2012)Van Der~Maaten and Weinberger]{van2012stochastic}
Van Der~Maaten, L. and Weinberger, K.
\newblock Stochastic triplet embedding.
\newblock In \emph{2012 IEEE International Workshop on Machine Learning for Signal Processing}, pp.\  1--6. IEEE, 2012.

\bibitem[Van Der~Maaten et~al.(2009)Van Der~Maaten, Postma, Van~den Herik, et~al.]{maaten2009dimensionality}
Van Der~Maaten, L., Postma, E., Van~den Herik, J., et~al.
\newblock Dimensionality reduction: a comparative.
\newblock Technical report, Tilburg University, 2009.

\bibitem[Wang et~al.(2021)Wang, Huang, Rudin, and Shaposhnik]{wang2021understanding}
Wang, Y., Huang, H., Rudin, C., and Shaposhnik, Y.
\newblock Understanding how dimension reduction tools work: an empirical approach to deciphering {t-SNE, UMAP, TriMAP, and PaCMAP} for data visualization.
\newblock \emph{Journal of Machine Learning Research}, 22\penalty0 (201):\penalty0 1--73, 2021.

\bibitem[Xiao et~al.(2017)Xiao, Rasul, and Vollgraf]{xiao2017fashion}
Xiao, H., Rasul, K., and Vollgraf, R.
\newblock {Fashion-MNIST}: a novel image dataset for benchmarking machine learning algorithms.
\newblock \emph{arXiv preprint arXiv:1708.07747}, 2017.

\bibitem[Zhou et~al.(2022)Zhou, Loy, and Dai]{zhou2022extract}
Zhou, C., Loy, C.~C., and Dai, B.
\newblock Extract free dense labels from {CLIP}.
\newblock In \emph{Proceedings of the European Conference on Computer Vision}, pp.\  696--712, 2022.

\bibitem[Zu \& Tao(2022)Zu and Tao]{zu2022spacemap}
Zu, X. and Tao, Q.
\newblock {SpaceMAP}: Visualizing high-dimensional data by space expansion.
\newblock In \emph{Proceedings of the International Conference on Machine Learning}, pp.\  27707--27723, 2022.

\end{thebibliography}
\bibliographystyle{icml2025}

\newpage
\appendix
\onecolumn
\section*{Appendix Summary}
This appendix presents a more detailed description of our method and the experimental results, which were omitted from the main text due to space limitations.
\par
Appendix~\ref{apx:att_rep} presents the attraction and repulsion computations used in this study. Although the resulting attraction force differs from that of the original paper~\citep{mcinnes2018umap}, we derive it by differentiating the objective function, and it is consistent with the original UMAP implementation. 
\par
Appendix~\ref{apx:exp_setting} presents the experimental setting in detail, including the computational resources and a description of the compared methods. We do not utilize special resources such as GPU, which motivates the applicability of our method.
\par
Appendix~\ref{apx:exp_results} presents additional experimental results. In the main text, we compared our method with popular baselines; however, other promising methods can be considered, including a parametric method~\citep{huang2024navigating}. In addition, the visualization results may be small and difficult to compare; thus, we present enlarged visualization results with additional comparisons.

\renewcommand{\theequation}{A.\arabic{equation}}
\setcounter{equation}{0}

\section{Analysis of Attraction and Repulsion of UMAP}
\label{apx:att_rep}
Here, we derive and analyze the attraction and repulsion of UMAP. First, we recall the loss function, i.e., the fuzzy cross-entropy loss:
\begin{align}
    \mathcal{L}&=-\sum_{i,j} w_{ij}\log v_{ij} - \sum_{i,j} (1-w_{ij})\log (1-v_{ij}),\label{apd_eq:obj}
\end{align}
where $w_{ij}$ is $(i,j)$ entry of the weighted adjacency matrix,  $ v_{ij}=\left(1+a||\bm{\mathrm{y}}_{i}-\bm{\mathrm{y}}_{j}||_{2}^{2b}\right)^{-1}$ with hyperparameters $a$ and $b$. Generally, $a\approx1.6$ and $b\approx0.9$. We then differentiate the loss with respect to $\bm{\mathrm{y}}_{i}$, which is the embedding to be optimized. First,let $n_{ij}=||\bm{\mathrm{y}}_{i}-\bm{\mathrm{y}}_{j}||_{2}^{2}$, and we differentiate $v_{ij}$ as follows:
\begin{align}
    \frac{\partial}{\partial \bm{\mathrm{y}}_{i}}v_{ij}&=\frac{\partial v_{ij}}{\partial n_{ij}}\frac{\partial n_{ij}}{\partial \bm{\mathrm
    y}_{i}}\notag\\
    &=-\frac{abn_{ij}^{b-1}}{(1+an_{ij}^{b})^{2}}\cdot2(\bm{\mathrm
    y}_{i}-\bm{\mathrm
    y}_{j})\notag\\
    &=-\frac{2ab||\bm{\mathrm{y}}_{i}-\bm{\mathrm{y}}_{j}||_{2}^{2(b-1)}}{\left\{1+a||\bm{\mathrm{y}}_{i}-\bm{\mathrm{y}}_{j}||_{2}^{2b}\right\}^{2}}(\bm{\mathrm{y}}_{i}-\bm{\mathrm{y}}_{j})\notag\\
    &=-2abv_{ij}^{2}||\bm{\mathrm{y}}_{i}-\bm{\mathrm{y}}_{j}||_{2}^{2(b-1)}(\bm{\mathrm{y}}_{i}-\bm{\mathrm{y}}_{j})\label{apd_eq:dv_dy}.
\end{align}
Then, we derive another expression of $\frac{1}{1-v_{ij}}$ as follows:
\begin{align}
    \frac{1}{1-v_{ij}}=\frac{1}{av_{ij}||\bm{\mathrm{y}}_{i}-\bm{\mathrm{y}}_{j}||_{2}^{2b}}.\label{apd_eq:1-v_inv}
\end{align}
Thus, the negative gradient of the loss function in Eq.~\eqref{apd_eq:obj} is expressed as follows:
\begin{align}
    -\frac{\partial}{\partial \bm{\mathrm{y}}_{i}}\mathcal{L}&=\sum_{j\neq i}w_{ij}\frac{\partial}{\partial \bm{\mathrm{y}}_{i}}\log v_{ij}+\sum_{j\neq i}(1-w_{ij})\frac{\partial}{\partial \bm{\mathrm{y}}_{i}}\log(1-v_{ij})\notag\\
    &=\sum_{j\neq i}w_{ij}\cdot\frac{1}{v_{ij}}\cdot(-2abv_{ij}^{2}||\bm{\mathrm{y}}_{i}-\bm{\mathrm{y}}_{j}||_{2}^{2(b-1)})(\bm{\mathrm{y}}_{i}-\bm{\mathrm{y}}_{j})\notag\\
    &\hspace{0.5cm}+\sum_{j\neq i}(1-w_{ij})\cdot\frac{1}{1-v_{ij}}\cdot2abv_{ij}^{2}||\bm{\mathrm{y}}_{i}-\bm{\mathrm{y}}_{j}||_{2}^{2(b-1)}(\bm{\mathrm{y}}_{i}-\bm{\mathrm{y}}_{j})\notag\\
    &=-\sum_{j\neq i}2ab||\bm{\mathrm{y}}_{i}-\bm{\mathrm{y}}_{j}||_{2}^{2(b-1)}v_{ij}w_{ij}(\bm{\mathrm{y}}_{i}-\bm{\mathrm{y}}_{j})+\sum_{j\neq i}\frac{2b}{||\bm{\mathrm{y}}_{i}-\bm{\mathrm{y}}_{j}||_{2}^{2}}v_{ij}(1-w_{ij})(\bm{\mathrm{y}}_{i}-\bm{\mathrm{y}}_{j})\notag\\
    &=\mathcal{A}_{i}+\mathcal{R}_{i}.
\end{align}
\par
\begin{figure*}
    \centering
    \includegraphics[width=160mm]{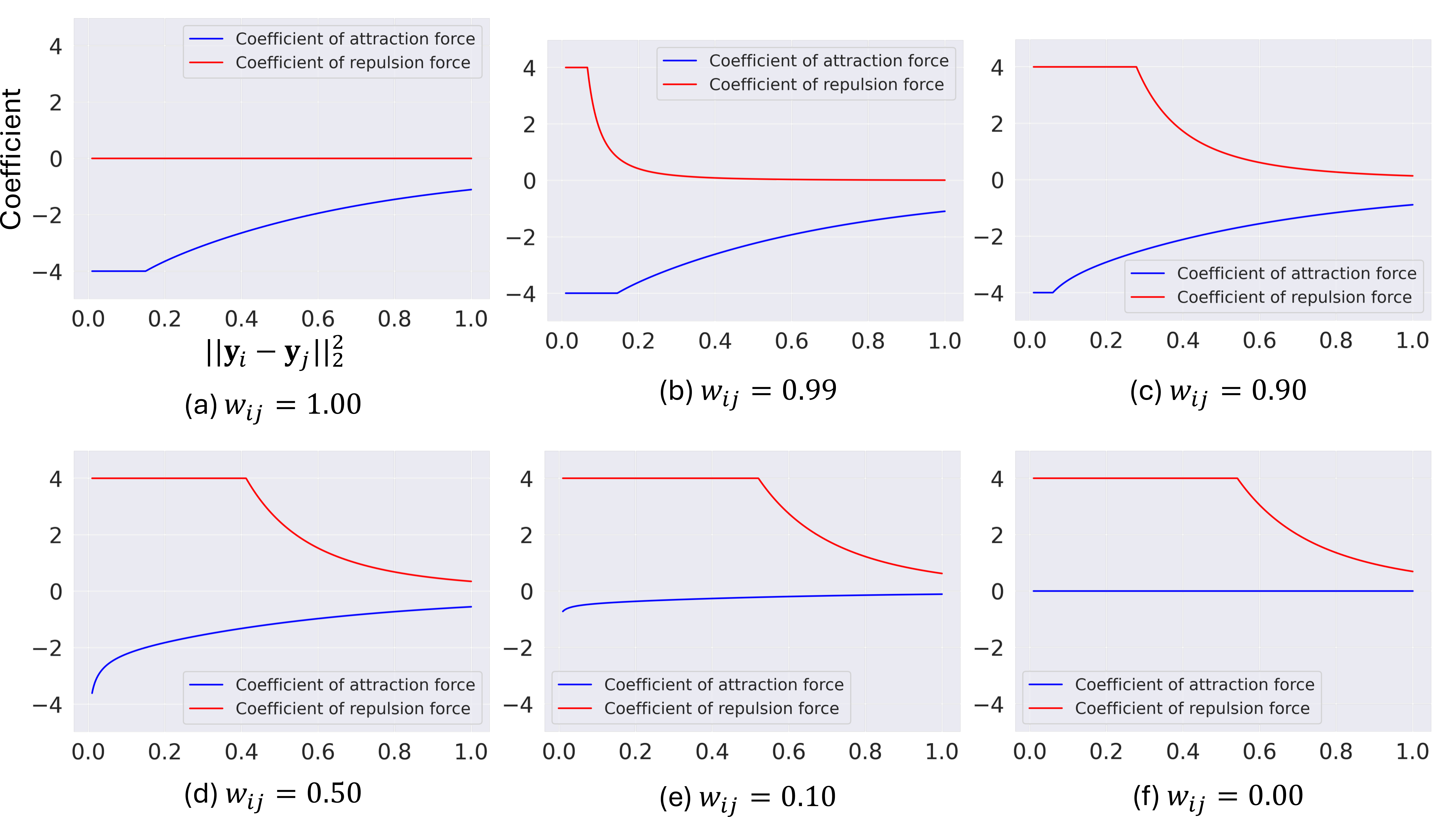}
    \caption{Comparison of coefficient between attraction $\mathcal{A}_{i}$ (\textcolor{blue}{blue}) and repulsion $\mathcal{R}_{i}$ (\textcolor{red}{red}) ($a=1.6,\,b=0.9$). The \textit{x-axis} is the distance between two embeddings $\bm{\mathrm{y}}_{i}$ and $\bm{\mathrm{y}}_{j}$ and the \textit{y-axis} is the coefficient of each force in Eq.~\eqref{eq:attraction} and Eq.~\eqref{eq:repulsion}. $w_{ij}$ is the similarity between two embeddings in the observed space, which should be preserved.}
    \label{fig:exp_vis_att_rep}
\end{figure*}

We then analyze the coefficients of attraction and repulsion. We denote the coefficient $a_{i}$ and $r_{i}$ as follows:
\begin{align}
    a_{i}&=-2ab||\bm{\mathrm{y}}_{i}-\bm{\mathrm{y}}_{j}||_{2}^{2(b-1)}v_{ij}w_{ij},\label{apd_eq:coef_attraction}\\
    r_{i}&=\frac{2b}{||\bm{\mathrm{y}}_{i}-\bm{\mathrm{y}}_{j}||_{2}^{2}}v_{ij}(1-w_{ij})\label{apd_eq:repulsion}.
\end{align}
The coefficients are clipped in $[-0.4,0.4]$ in the practical UMAP implementation. Note that, $a_{i}$ and $r_{i}$ depend on the distance $d_{ij}=||\bm{\mathrm{y}}_{i}-\bm{\mathrm{y}}_{j}||_{2}$. We plot them as the functions $a_{i}(d_{ij})$ and $r_{i}(d_{ij})$ with a constant $w_{ij}$ in Figure~\ref{fig:exp_vis_att_rep}. When $w_{ij}$ is large, $a_{i}$ obtains large values, and $r_{i}$ obtains small values. This implies that two embeddings with high similarities get closer. In contrast, when $w_{ij}$ is small, $a_{i}$ obtains small values, and $r_{i}$ obtains large values, which implies that two embeddings with small similarities become distant. This property allows $\mathcal{A}_{i}$ and $\mathcal{R}_{i}$ to be considered \textit{attraction} and \textit{repulsion}, respectively. In addition, the behavior of the attraction and repulsion forces implies the importance of the initial embedding, i.e., the attraction and repulsion forces do not act on the distinct points (e.g., $d_{ij}\geq1$). Generally, the absolute value of $r_{i}$ is greater than that of $a_{i}$, which indicates repulsion is stronger than attraction, and UMAP tends to separate the dissimilar points.

\section{Detailed Experimental Settings}
\label{apx:exp_setting}
In this study, our experiment was conducted on an Intel Core i7-10700 CPU with 16GB RAM. In addition, we used the umap-learn library to implement our method and partially used the code in the literature~\citep{klimovskaia2020poincare} to produce the visualization. Note that we did not vary $\lambda$ and $k$, i.e., these parameters were fixed $\lambda=0.1$ and $k=20$. In contrast, we varied $C$ across the dataset and recommended a relatively high value ($C\approx100$) as the default setting. 
\par
The following $5$ methods were compared in the experimental evaluations.
\begin{enumerate}[label=\Roman*., topsep=0.0em, itemsep=-0.1em]
   \item\textbf{PCA}. PCA is a popular method that embeds data based on its singular vectors. In this study, we used the scikit-learn implementation~\citep{pedregosa2011scikit} of PCA.
  \item\textbf{t-SNE}~\citep{van2012stochastic}. t-SNE is a gold standard method for data visualization. Note that the original t-SNE algorithm is slow and does not scale data with $N>10,000$; thus, in this study, we employed the Barnes-Hut approximation method~\citep{van2014accelerating} in the scikit-learn library~\citep{pedregosa2011scikit}.
  \item\textbf{PHATE}~\citep{moon2019visualizing}. PHATE is based on diffusion maps~\citep{coifman2006diffusion} and multidimensional scaling~\citep{torgerson1958theory}. PHATE estimates the multihop neighbor relation, which enables neighbor-based global preservation. In this study, we used the implementation provided by the original paper.
  \item\textbf{UMAP}~\citep{mcinnes2018umap}. UMAP is also a widely used neighbor embedding method that is generally faster than t-SNE. Here, we used the original implementation.
  \item\textbf{PaCMAP}~\citep{wang2021understanding}. PaCMAP introduces the middle-neighbor relation into the UMAP algorithm and well defines the initialization value. In this study, we used its open-source implementation presented in the original paper.
\end{enumerate}

\section{Additional Experimental Results}
\label{apx:exp_results}

\subsection{Additional Results for the MNIST Dataset}
\begin{figure*}
    \centering
    \includegraphics[width=150mm]{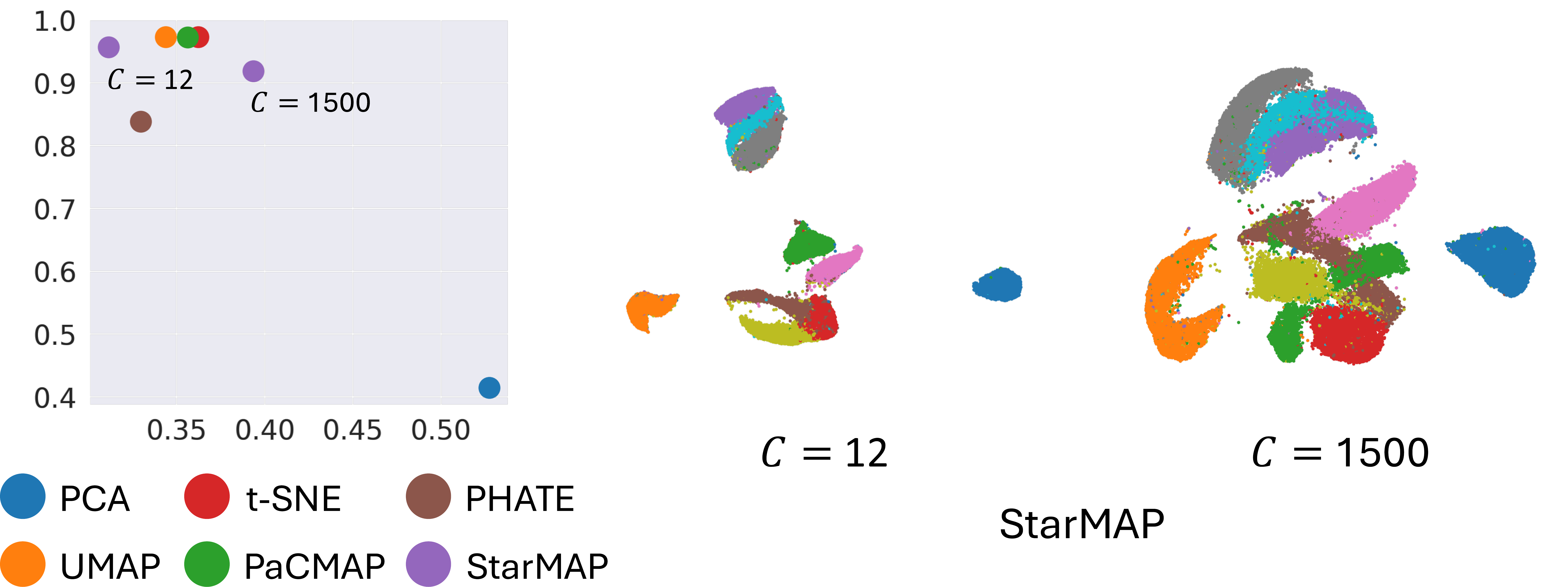}
    \caption{Quantitaive (\textit{left}) and qualitative results (\textit{left}) for the MNIST dataset with different StarMAP settings.}
    \label{fig:exp_mnist_eval}
\end{figure*}

In the experiments discussed in the main paper, StarMAP was inferior to the neighbor embedding methods on the MNIST dataset. However, StarMAP with different hyperparameters obtained a higher distance correlation score (Figure~\ref{fig:exp_mnist_eval}). However, the local accuracy degraded, and the cluster corresponding to the digit `2' split in the visualization with a large $C$ value. These results suggest one hypothesis concerning the MNIST dataset, i.e., the digit clusters exhibited in the MNIST dataset are complicated and originally separated. The identification of clusters inherently complicates the preservation of the global structure. As clustering emphasizes local relationships and discriminative separation, it can distort or overshadow broader structural similarities within the data. Balancing local cluster discrimination while maintaining the global data structure remains a key challenge in dimensionality reduction and visualization techniques.
\label{apx:mnist}

\subsection{Visualization of the Star Position}
\label{apx:star_vis}
Next, we visualize the star position in StarMAP. Stars are the anchor embedding and have the attraction force to their points in the observed space. Here, we expected them to retain the global structure based on the PCA embedding.
\par
\begin{figure*}
    \centering
    \includegraphics[width=150mm]{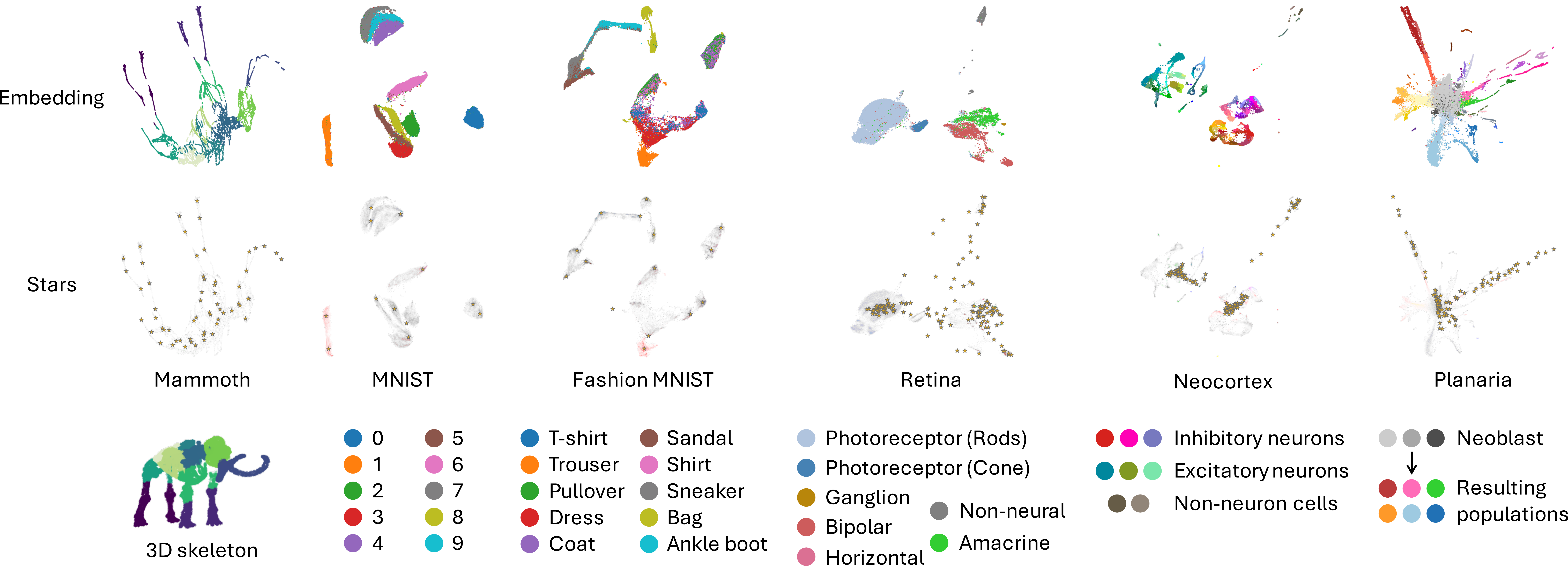}
    \caption{Visualization of the star position on six real-world datasets. The color codes of each cluster are denoted below the visualizations.}
    \label{fig:exp_visualizing_stars}
\end{figure*}

Figure~\ref{fig:exp_visualizing_stars} shows the star visualizations obtained on the six datasets. For the Mammoth dataset, the stars well preserved the outline of the skeleton, and for the MNIST, Fashion MNIST, and Retina datasets, stars were assigned to each cluster, enabling preservation of the intercluster similarity. However, on the Cortex and Planaria datasets, the stars were not assigned to each cluster and were less informative about the cluster information of the datasets. StarMAP embeds data with the star and neighbor attraction, and even on complicated datasets, with stars that are less informative about the local clusters, StarMAP embeds the data faithfully using the neighbor information. 

\subsection{Heuristic Implementation of StarMAP}
\label{apx:heuristic}
We extend StarMAP to more practical use cases and confirm its behavior. First, we introduce the automatic determination of $C$. As mentioned in the main paper, the automatic determination method of $K$-means clustering is generally sensitive to noise. Thus, we determine it proportionally to the sample size as $C=\min(\frac{N}{500}, 100)$. Second, we perform PCA if $D>50$ before $K$-means clustering. This reduces the time complexity linearly and accelerates the entire algorithm. We compared this method with UMAP on six datasets (Section~\ref{subsec:real_world_data_vis}), where we evaluated them based on the average time over 10 runs and their visualization results.
\par
\begin{figure*}[t]
    \centering
    \includegraphics[width=150mm]{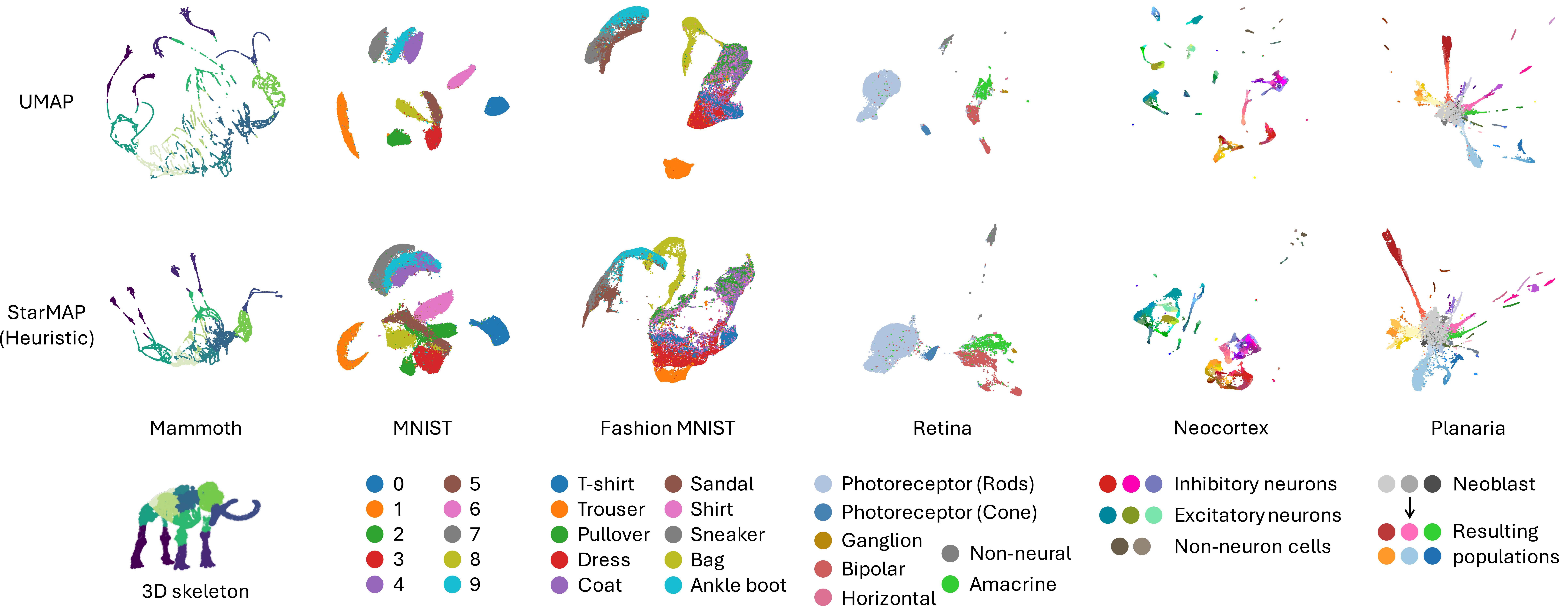}
    \caption{Visualization results of UMAP and StarMAP with heuristic implementation.}
    \label{fig:exp_heuristic_visualization}
\end{figure*}

Table~\ref{tab:time_results} shows the average run time compared to UMAP. Although StarMAP was slower than UMAP, the average runtime results are acceptable. The implementation elaboration may improve the run time e.g., using GPU acceleration of $K$-means clustering. Figure~\ref{fig:exp_heuristic_visualization} shows the corresponding visualization results. As can be seen, StarMAP produced visualizations that are similar to the manually selected version except for the MNIST dataset. 
\begin{table*}[t]
\caption{Time comparison between StarMAP and UMAP. The time is averaged over 10 runs.}
  \begin{center}
      \begin{tabular}{lccccccc} \toprule
         \textit{Datasets} & Mammoth & MNIST & FMNIST & Retina & Neocortex & Planaria \\ \midrule
         UMAP & 36.6 & 126.1 & 131.6 & 106.9 & 57.3 & 48.9 \\
         StarMAP (Heuristic) & 46.6 & 156.2 & 159.5 & 125.0 & 69.8 & 57.9 \\
        \bottomrule
      \end{tabular}
    \label{tab:time_results}
  \end{center}
\end{table*}

\subsection{Visualization Comparison with Additional Methods}
\label{apx:vis_comparison}
\begin{figure*}
    \centering
    \includegraphics[width=150mm]{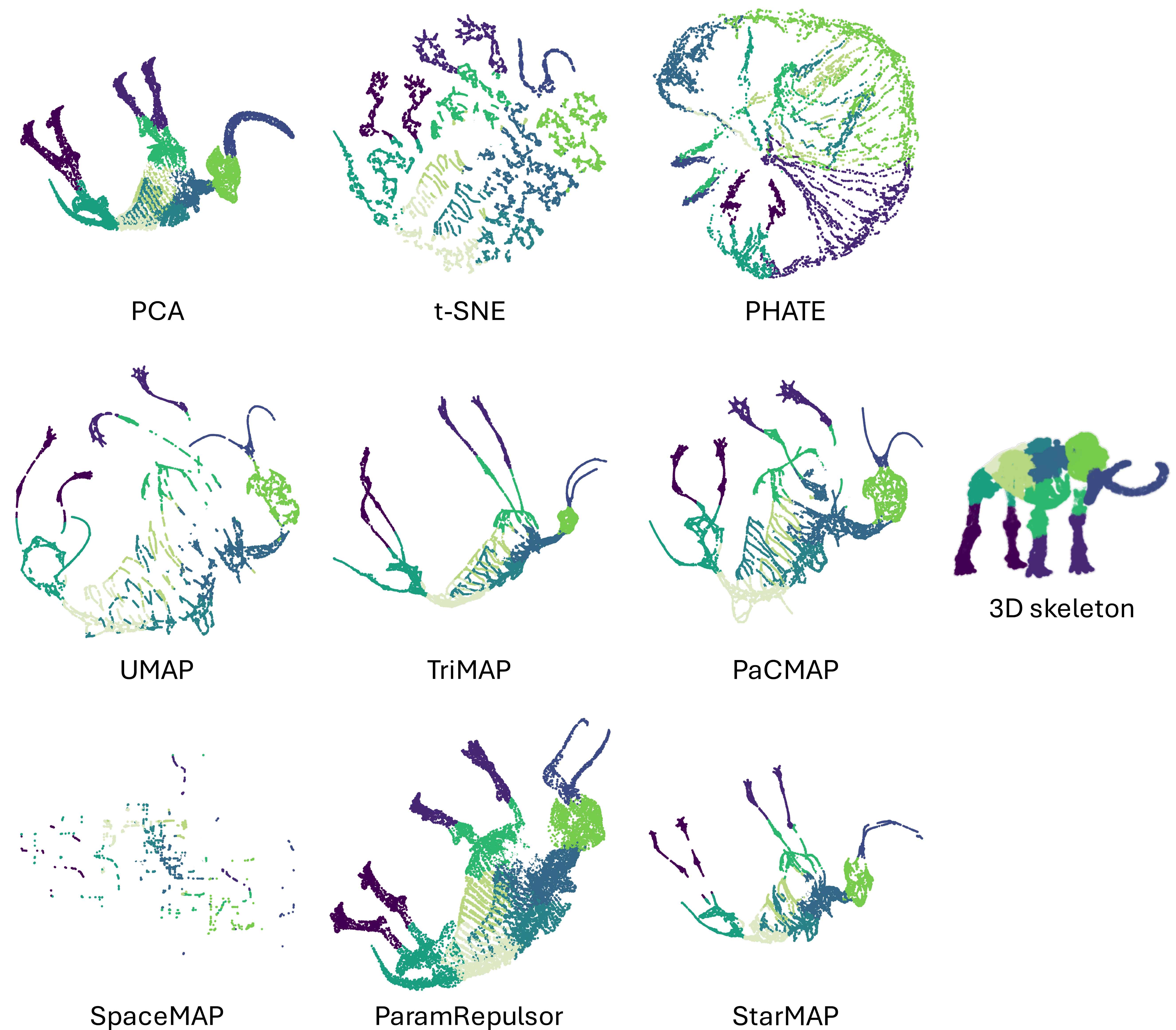}
    \caption{Visualization results on Mammoth dataset.}
    \label{fig:exp_mammoth_all}
\end{figure*}

\begin{figure*}
    \centering
    \includegraphics[width=150mm]{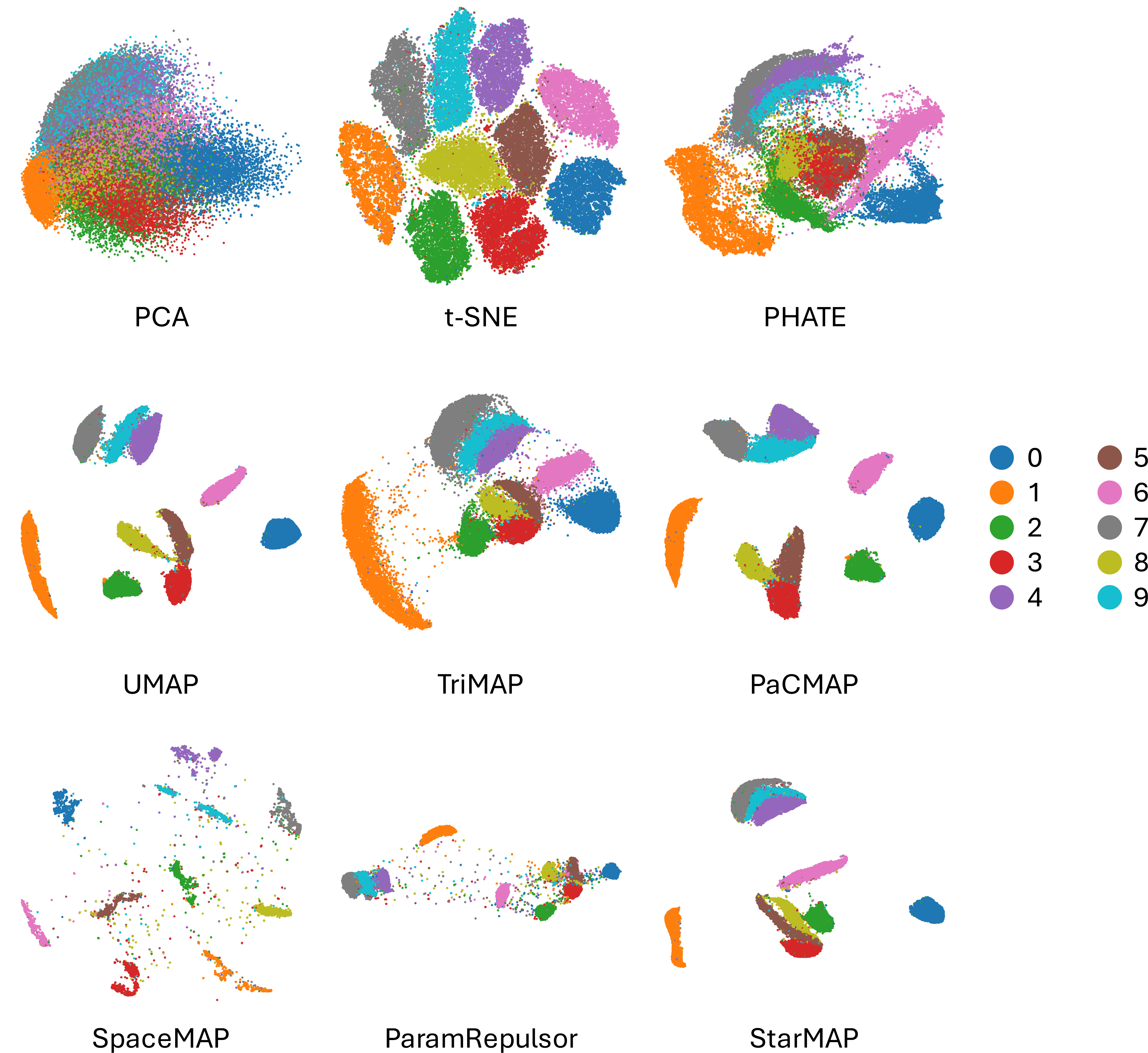}
    \caption{Visualization results on MNIST dataset.}
    \label{fig:exp_mnist_all}
\end{figure*}

\begin{figure*}
    \centering
    \includegraphics[width=150mm]{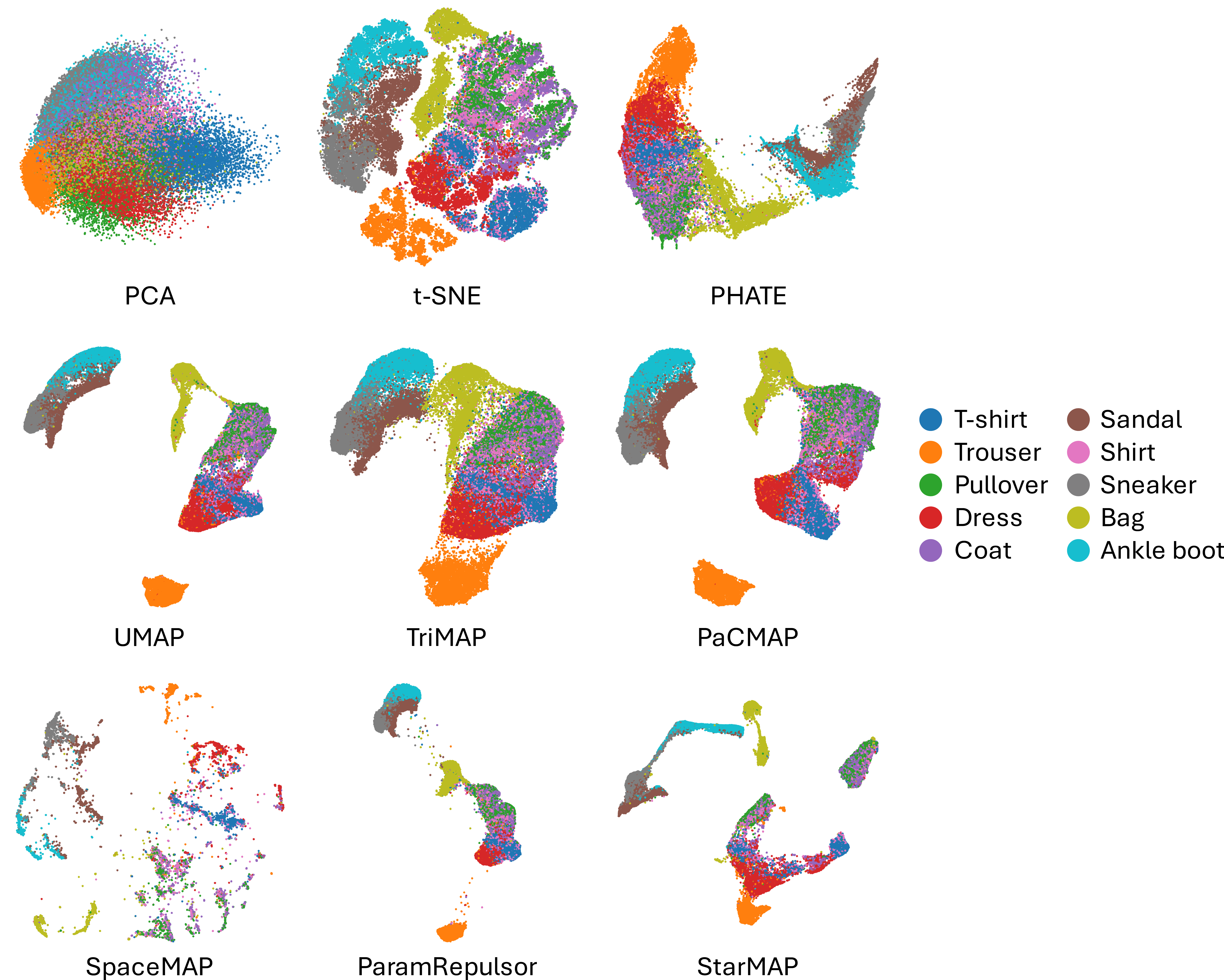}
    \caption{Visualization results on Fashion MNIST dataset.}
    \label{fig:exp_fmnist_all}
\end{figure*}

\begin{figure*}
    \centering
    \includegraphics[width=150mm]{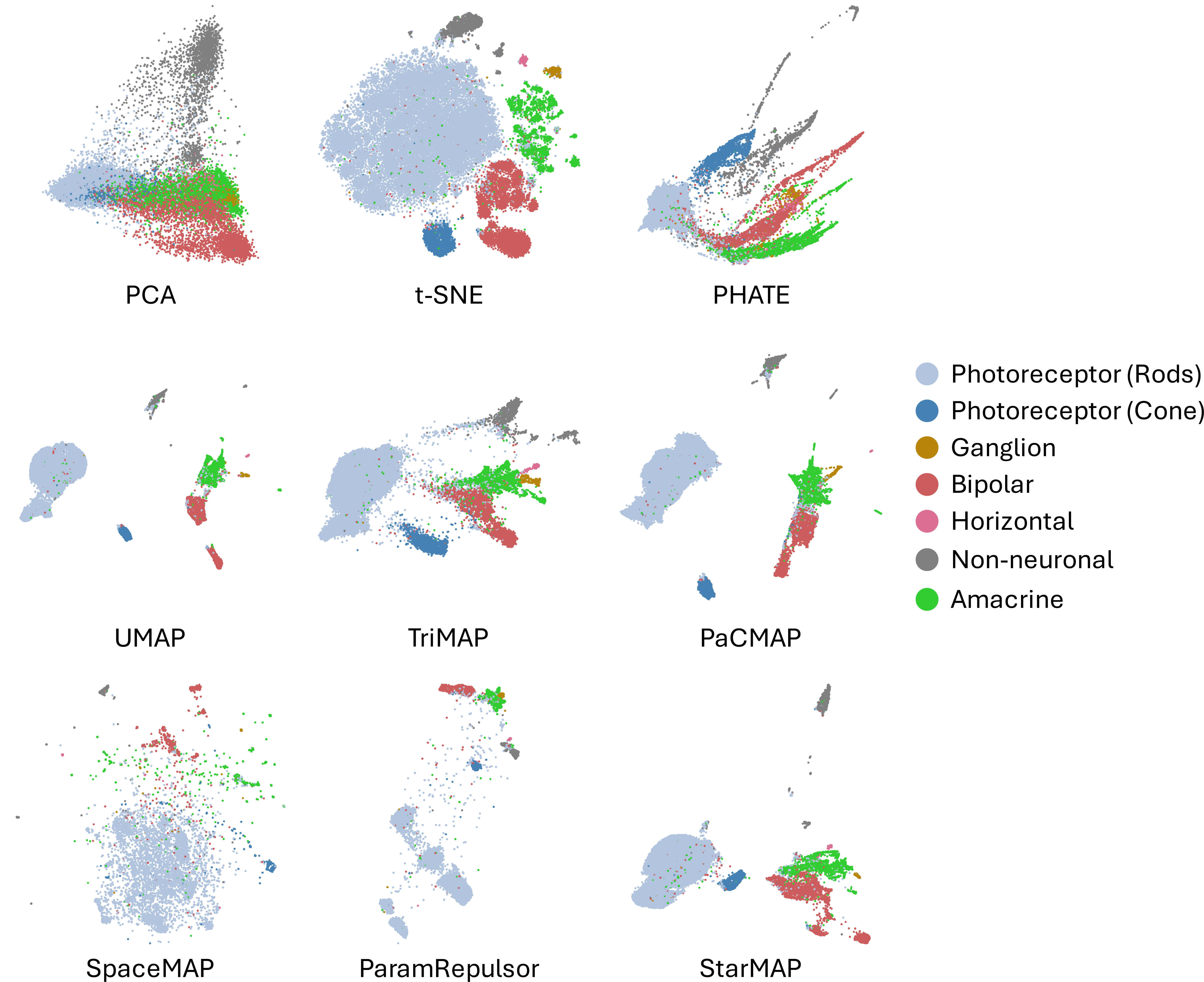}
    \caption{Visualization results on Retina dataset.}
    \label{fig:exp_macosko_all}
\end{figure*}

\begin{figure*}
    \centering
    \includegraphics[width=150mm]{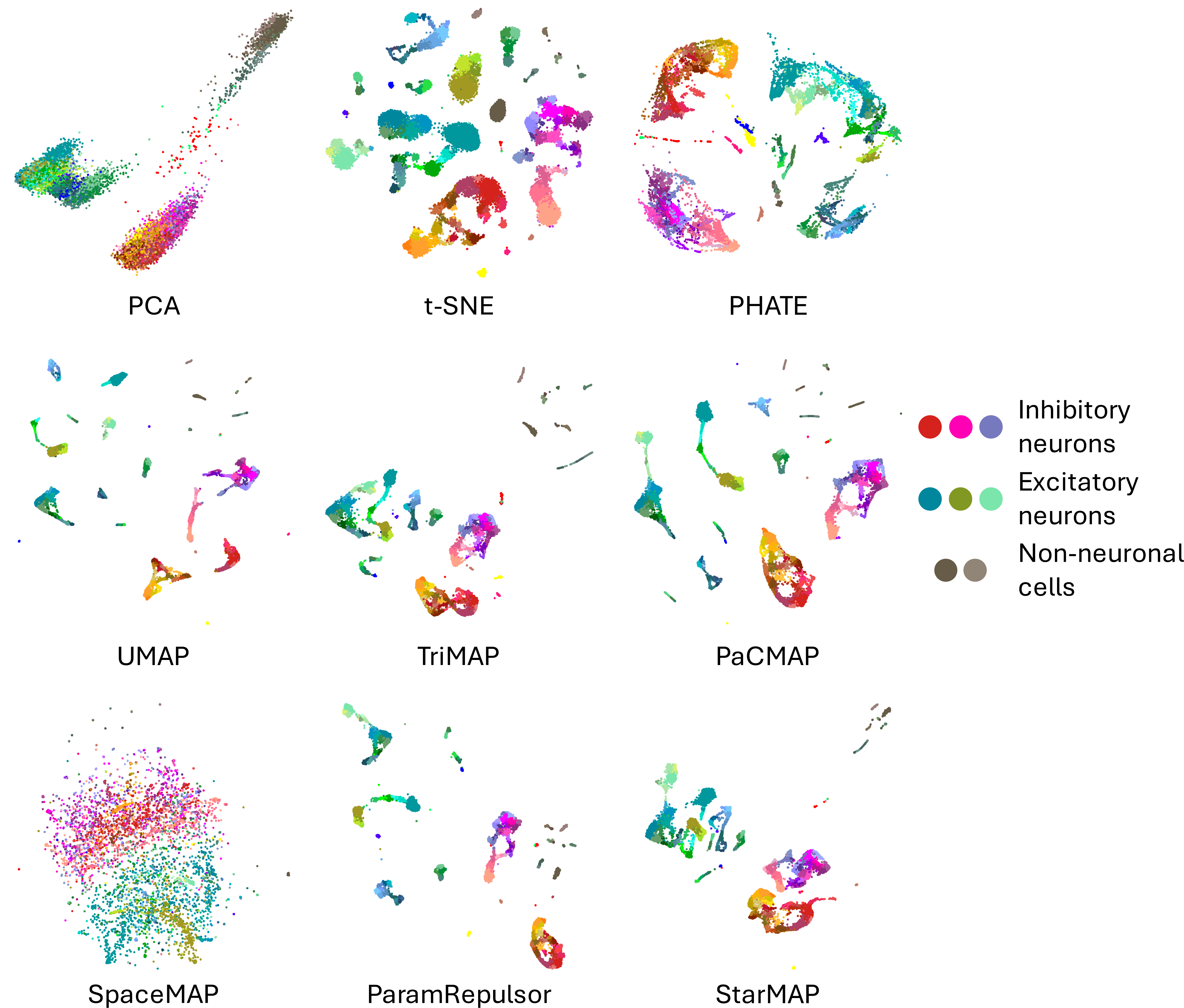}
    \caption{Visualization results on Cortex dataset.}
    \label{fig:exp_tasic_all}
\end{figure*}

\begin{figure*}
    \centering
    \includegraphics[width=150mm]{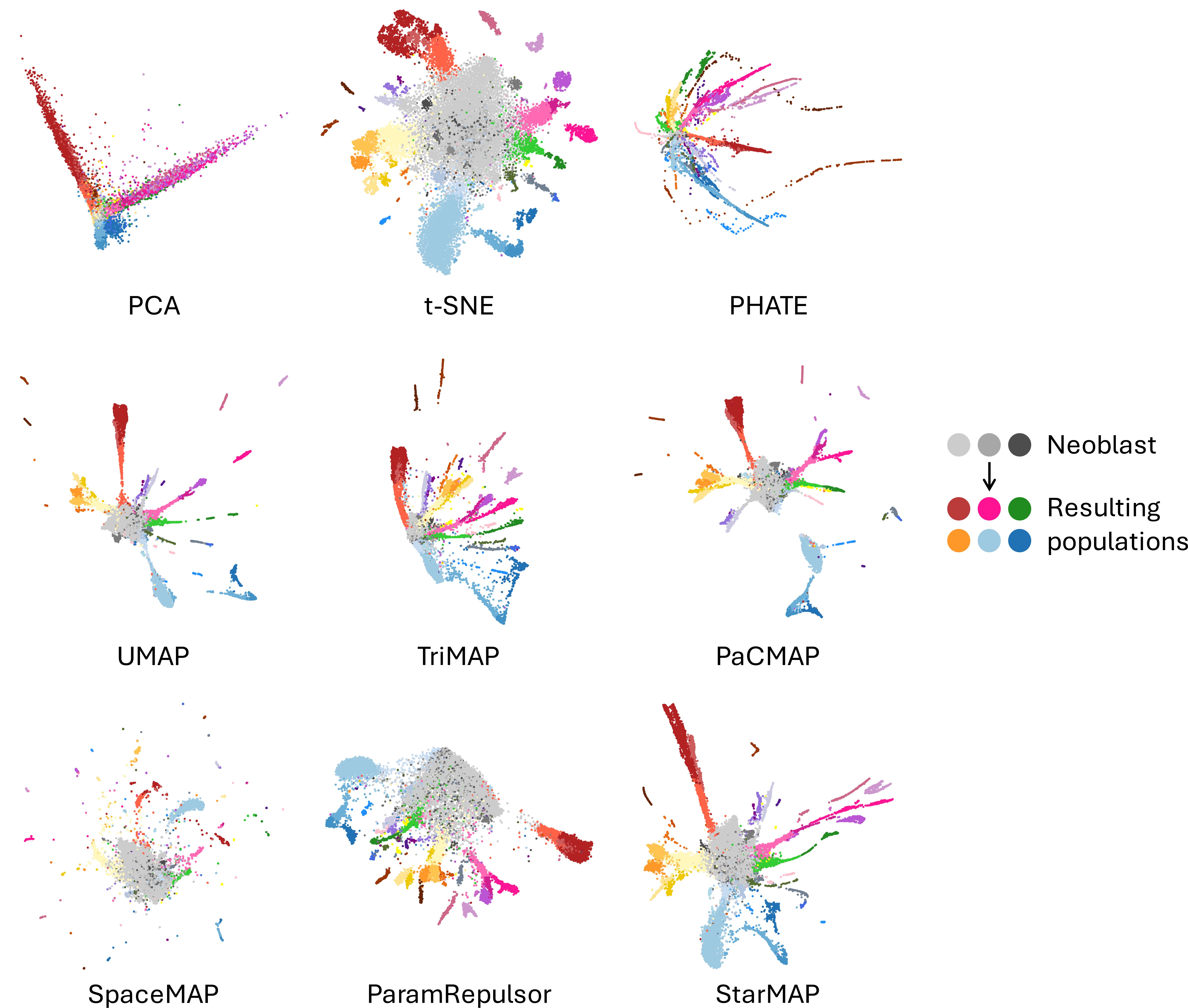}
    \caption{Visualization results on Planaria dataset.}
    \label{fig:exp_planaria_all}
\end{figure*}

We also compared StarMAP with the following $3$ methods.
\begin{enumerate}[label=\Roman*., topsep=0.0em, itemsep=-0.1em]
   \item\textbf{TriMAP}~\citep{amid2019trimap}. TriMAP introduces the triplet relation to embed data and has approximately the same complexity as UMAP. In this study, we used the implementation presented in the original paper.
   \item\textbf{SpaceMAP}~\citep{zu2022spacemap}. SpaceMAP is based on space expansion and considers the discrepancy of geometry between high- and low-dimensional space. We also used its open-source implementation presented in the original paper. 
    \item\textbf{ParamRepulsor}~\citep{huang2024navigating}. ParamRepulsor is a parametric neighbor embedding method that has the parametric mapping from high-dimensional into low-dimensional space. The implementation was used in the current study.
\end{enumerate}
Figure~\ref{fig:exp_mammoth_all} shows the error bar plot of all methods. After several promising compared methods added, StarMAP still appeared in the upper-right corner, validating its effectiveness. Figure~\ref{fig:exp_mammoth_all} shows the Mammoth dataset visualization. As can be seen, PCA, ParamRepulsor, and StarMAP preserved the outline of the mammoth. Figure~\ref{fig:exp_mnist_all} shows the MNIST visualization. t-SNE, UMAP, TriMAP, PaCMAP, and StarMAP preserved the digit clusters. Although SpaceMAP and ParamRepulsor recognized each cluster, they contained cluttered points and were less visible. Figure~\ref{fig:exp_fmnist_all} shows the Fashion MNIST dataset visualization, which demonstrated similar tendencies as the MNIST visualization. Figure~\ref{fig:exp_macosko_all} shows the Retina dataset visualization. Although all of the methods (except PCA and PHATE) recognized the photoreceptor, bipolar, non-neuronal, and amacrine populations, t-SNE, TriMAP, UMAP, and StarMAP only contained the horizontal cell clusters. Figure~\ref{fig:exp_tasic_all} presents the Retina dataset visualization. TriMAP and StarMAP preserved the cell clusters and inter-population similarity. Note that, TriMAP employs triplet similarity, which may be beneficial with this dataset, which has the three largest cell groups. Finally, Figure~\ref{fig:exp_planaria_all} presents the Planaria dataset visualization. Although PCA, PHATE, and StarMAP retained the lineage along the differentiation, StarMAP preserved the neoblast and resulting population clusters most efficiently. 

\begin{figure*}
    \centering
    \includegraphics[width=170mm]{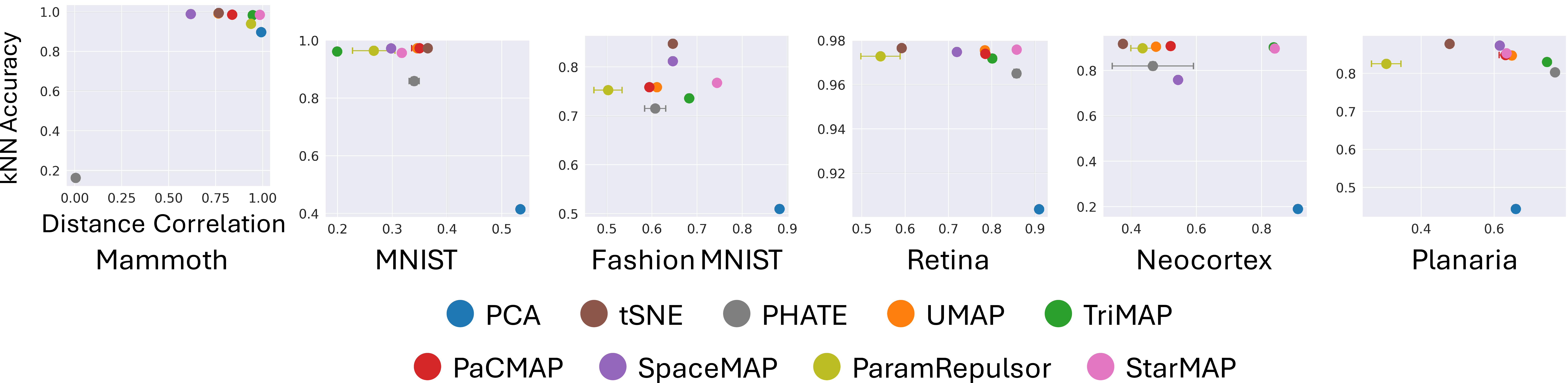}
    \caption{Quantitative results of error plots with distance correlation (\textit{global}, \textit{x-axis}) and $k$NN accuracy (\textit{local}, \textit{y-axis}).}
    \label{fig:exp_error_bar_all}
\end{figure*}

\end{document}